%% file: main.tex
\documentclass[a4paper]{article}
\usepackage[utf8]{inputenc}
\usepackage{graphicx}
\usepackage{times}
\usepackage{float}
\usepackage{paralist}
\usepackage{todonotes}
\usepackage{hyperref}
\hypersetup{
	colorlinks=true,
	linkcolor=black,
	filecolor=black,      
	urlcolor=black,
	citecolor=blue,
}
\usepackage{soul}
\usepackage{smartdiagram}
\usepackage{pgf}
\usepackage{tikz}
\usepackage{pgfplots}
\usetikzlibrary{arrows,decorations,backgrounds,matrix,automata,
	trees,shapes,shadows,plotmarks,calc,positioning,patterns,chains,fit}
\usepgfplotslibrary{patchplots,groupplots}
\pgfplotsset{compat=1.16}

\usepackage{xspace}
\usepackage[linesnumbered,commentsnumbered,ruled]{algorithm2e}
\usepackage{amssymb}
\usepackage{natbib}
\usepackage{paralist}
\usepackage{multirow}
\usepackage{authblk}

\usepackage[acronym]{glossaries}
\newacronym{osm}{OSM}{Open Street Map}
\newacronym{dem}{DEM}{Digital Elevation Model}
\newacronym{dtm}{DTM}{Digital Terrain Model}
\newacronym{gtfs}{GTFS}{General Transit Feed Specification}
\newacronym{matsim}{MATSim}{Multi-Agent Transport Simulation}
\newacronym{ivabm}{IV-ABM}{Infrastructure Victoria Activity Based Model}
\newacronym{abm}{ABM}{Activity-Based Model}
\newacronym{vitm}{VITM}{Victorian Integrated Transport Model}
\newacronym{vista}{VISTA}{Victorian Integrated Survey for Travel and Activity}
\newacronym{abs}{ABS}{Australian Bureau of Statistics}
\newacronym{pt}{PT}{Public Transportation}
\newacronym{csv}{CSV}{Comma Separated Values}
\newacronym{sa1}{SA1}{Statistical Area level 1}
\newacronym{sa2}{SA2}{Statistical Area level 2}
\newacronym{sa3}{SA3}{Statistical Area level 3}
\newacronym{sa4}{SA4}{Statistical Area level 4}
\newacronym{tasha}{TASHA}{Travel Activity Scheduler for Household Agents}
\newacronym{cbd}{CBD}{Central Business District}

\newcommand{\kk}[1]{\textsf{\footnotesize{#1}}}

\newcommand{\vista}{\acrshort{vista}\xspace}
\newcommand{\csv}{\acrshort{csv}\xspace}

\newcommand{\A}{$\mathcal{A}$\xspace}
\newcommand{\C}{$\mathcal{C}$\xspace}
\newcommand{\D}{$\mathcal{D}$\xspace}
\newcommand{\Ds}{$\mathcal{D}_{s}$\xspace}
\newcommand{\Dso}{$\acute{\mathcal{D}_{s}}$\xspace}
\newcommand{\Deo}{$\acute{\mathcal{D}_{e}}$\xspace}
\newcommand{\De}{$\mathcal{D}_{e}$\xspace}
\newcommand{\N}{$\mathcal{N}$\xspace}
\newcommand{\T}{$\mathcal{T}$\xspace}

\newcommand{\Omit}[1]{}

\def\zz#1#2{\small\begin{tabular}[t]{l}\hline#1\\\hline\hline#2\end{tabular}\linebreak[0]\ignorespaces}

\title{An Activity-Based Model of Transport Demand for Greater Melbourne}

\author[a]{Alan Both}
\author[b,c]{Dhirendra Singh}
\author[a]{Afshin Jafari\thanks{Corresponding Author. Email address: afshin.jafari@rmit.edu.au. Postal address: RMIT University, GPO Box 2476, Melbourne VIC 3001.}}
\author[a]{Billie Giles-Corti}
\author[a]{Lucy Gunn}

\affil[a]{School of Global, Urban and Social Studies, RMIT University}
\affil[b]{School of Computing Technologies, RMIT University}
\affil[c]{Data61, CSIRO}

\date{}

\begin{document}
	
	\maketitle
	%\listoftodos
	
	%\newpage
	
	\begin{abstract}
		
		In this paper, we present an activity-based model for the Greater Melbourne area, using a combination of hierarchical clustering, probabilistic, and gravity-based approaches. The model outlines steps for generating a synthetic population—a list of agents with their demographic attributes—and for assigning activity patterns, schedules, as well as activity locations and modes of travel for each trip.
		
		In our model, individuals are assigned activity chains based on the probabilities of their respective demographic clusters, as informed by observed data. Tours and trips then emanate from these assigned activities. This is innovative compared to the common practice of creating trips or tours first and attaching activities thereafter. Furthermore, when selecting activity locations, our model incorporates both the distance-decay of trip lengths and the activity-based attraction of destination sites. This results in areas with higher attractiveness for various activities showing a greater likelihood of being selected. Additionally, when assigning the location for the next activity, we take into account the number of activities an agent has remaining to ensure they do not opt for a location that would be impractical for a return trip home.
		
		Our methodology is open and replicable, requiring only publicly available data and is designed to produce outcomes compatible with commonly used agent-based modeling software such as MATSim. Each sub-model is calibrated to match observed data in terms of activity types, start and end times, and durations.
		
	\end{abstract}
	

	\section{Introduction} %%%%%%%%%%%%%%%%%%%%%%%%%%%%%%%%%%%%%%%%%%%%%%%%%%%%%%%%%%%
	\label{sec_introduction}
	
	\glspl{abm} have been extensively used in both private and public sectors to simulate city-wide transportation systems \citep{milakis_what_2014,infrastructure_victoria_automated_2018,zhang2018integrated,bekhor2011integration,knapen2021activity,kpmg__arup_model_2017}.
	In these examples, the travel behaviors of individual travelers are studied within an \gls{abm} simulation to assess the impact of policies and test scenarios on mode choices, travel itineraries and traffic flows.
	In doing so, these models provide much-needed evidence to understand transport systems and land uses and to fine-tune policies to better support planning and decision making \citep{miller2021activity}, prior to the implementation of costly infrastructure changes. 
	\glspl{abm} can also be used to evaluate competing policies, for example, in reducing congestion and moderating traffic flows. These are major challenges that are recognized globally by transport planners and governments as impacting the livability and sustainability of growing cities \citep{victoria_state_government_plan_2014,auckland_council_auckland_2018,city_of_toronto_official_2015,city_of_portland_b_o_p_a_s_portland_2009}.
	In this sense, \glspl{abm} provide policymakers with a virtual laboratory to enhance their decision-making.
	
	When using an \gls{abm} to simulate individual-level travel behavior, particularly when including active modes of transport such as walking and cycling, it is necessary for the synthetic travelers to be assigned individual-level demographic information such as age \citep{chang2013factors,haustein2012mobility}, sex \citep{cheng2017improving}, household characteristics such as income \citep{ko2019exploring,cui2019accessibility,allen2020planning}, or the presence of children \citep{ o2004constraints} since these attributes are associated with transportation mode choices and consequently travel behavior \citep{ha2020unraveling,ding2017exploring,manaugh2015importance,cervero2002built}.
	Some simulation models include attributes such as car ownership or access \citep{liu_development_2020,scherr_towards_2020}, income to model the impact of fuel prices; possession of a concession card \citep{infrastructure_victoria_automated_2018}; or whether a traveler delivers something or drops someone off \citep{horl2020open}.
	In an activity-based modeling environment, it is therefore important that key components of travel behavior are included in detail, such as individual demographic attributes and features of the home- and work-related environments in which individuals circulate.
	
	Information on trip timing and stops or the activities that individuals undertake can be obtained from travel survey diaries; it is nonetheless necessary to develop a process that does not replicate or clone travelers from an existing sample of real-life individuals, but instead generates new versions of them according to demographic, activity-based, location, and trip attributes that are most likely to be present in an area, given the underlying survey or population data from which the agents were derived.
	For clarity, we refer to these new synthetic travelers as agents, the collection of these agents matching the census as the synthetic population, and the list of agents and their daily activities and travel diaries as the activity-based transport demand. 
	
	One issue in creating synthetic population-based transport demand models is that data on individuals and their travel behaviors can be expensive to collect, or, when data exist, may be aggregated or anonymized to protect the privacy of the individuals.
	To overcome these problems, various techniques have been developed to create a synthetic population with demographic attributes from limited existing data sources and to assign travel details to them.
	\citet{wang_improved_2021} divided the process into generating agents with demographics, which we refer to as the synthetic population generation step, assigning activity patterns, including activity chains and timing, and assigning locations to activities. 
	
	To generate the \textbf{\textit{synthetic population}}, \cite{rahman_methodological_2010} classified approaches to generate synthetic agents with demographic attributes into two main categories: synthetic reconstruction and re-weighting, re-weighting being the more recent category \citep{hermes_review_2012}.
	Synthetic reconstruction typically uses a list of agents and combines their basic demographic attributes with data on home location derived from sources such as a census. Additional demographic attributes of interest are added to this initial list based on conditional probabilities and a sequential attribute addition process \citep{williamson_evaluation_2013}.
	In re-weighting, rather than creating synthetic individuals, each observation from the travel survey is assigned a weight indicating how representative that observation is of each area.
	For example, an observation might represent multiple individuals in one area and no one in another area.
	These weights are calculated and adjusted so that the distribution of the synthetic population reflects that of the observed data \citep{williamson_evaluation_2013,hermes_review_2012}.
	
	Central to the creation of any \gls{abm} is the \textbf{\textit{assignment of activity chains}}, where each agent is assigned a series of activities related to their travel behavior and timing for each activity (that is, its start time and duration). These activities then form the origin and destination of a trip \citep{wang_improved_2021,lum_two-stage_2016}, in sequence forming a chain.
	While there are multiple approaches for the generation of activity chains that reflect traveller choices, most fall under either a conditional approach, where a set of conditional probabilities based on attributes of travelers, such as occupation \citep{he_evaluation_2020} or demographic attributes \citep{balac_synthetic_2021} are used; an approach based on statistical models, such as CEMDAP \citep{bhat2004comprehensive}; or finally cloning-based approaches where activity chains are randomly selected from existing data \citep{felbermair_generating_2020}.
	A major benefit of the conditional and statistical approaches is that it allows for scenario-based adjustment of travel behavior prior to the running of network assignment models such as MATSim, but at the cost of being restricted to tour-based activity chains, where activities are assigned to tour components using a nested logit function.
	There are however other approaches, such as activity scheduling models, in development.
	Similar to cloning-based approaches, activity scheduling models lack the inherent rigidity of tour-based models, but obtain far more flexibility and variety by assigning activities according to type, start time, and duration without the use of a tour-based paradigm \citep{nayak_critical_2022,miller_current_2023}.
	This lack of a higher level structure constraining the activities allows tours to instead form organically, although they currently lack the ability to apply scenario-based adjustment of travel behavior prior to the running of network assignment.
	
	For \textbf{\textit{activity location assignment}}, gravity models are commonly used.
	Gravity models select activity locations according to an inversely proportional distance from the origin or anchor locations (e.g., home / work), along with origin and destination matrices and random assignment \citep{lum_two-stage_2016}.
	\cite{nurul_habib_comprehensive_2018} proposed a model in which activities, timings, and locations were jointly assigned based on random utility maximization theory. 
	
	Following the process of generating the synthetic population-based transport demand discussed above, \cite{sallard_synthetic_2020} generated a transport demand model for the city of Sao Paulo.
	For home-related activities, they assigned a random residential location to each household.
	For work-related activities, the location assignment was based on the origin–destination work trip counts with travel distances extracted from a travel survey.
	They divided education trips into different groups according to home location, sex, and age of each survey respondent.
	The location of the education destination for each agent was then assigned based on the trip distance density function related to each respective group.
	Finally, secondary activity locations (e.g., leisure, shopping, other) were assigned using a gravity model based on realistic travel distances to these destinations.
	A similar process was followed in \cite{balac_synthetic_2021} to assign secondary activity locations.
	\cite{ziemke_matsim_2019} used the econometric model, CEMDAP, to create activity patterns and an initial location assignment, which then used the MATSim agent-based traffic simulation toolkit to adjust the locations assigned to ensure that the resulting traffic matched the observed data best.
	
	Another widely used travel demand and schedule generator, \gls{tasha}, was developed by \cite{roorda_validation_2008}.
	They used demographics from the Greater Toronto travel survey to develop joint probability functions for activity type, demographics, household structure, and trip schedules.
	An additional probabilistic approach was applied to select the time and duration for each activity.
	The resulting 262 distributions were used to generate activity chains for each individual.
	The inputs into \gls{tasha} include home and work locations, while other activities were assigned using entropy models based on distance, employment and population density and land use measures such as floor space of the shopping center \citep{roorda_validation_2008}.
	A more recent approach that has improved synthetic population-based transport demand generation with greater accuracy and flexibility is Machine Learning \citep{koushik2020machine}.
	Using a hybrid framework, \citet{hesam2021framework} combined machine learning with econometric techniques to create activity chains and travel diary using a cohort-based synthetic pseudo-panel engine.
	Similarly, \cite{allahviranloo2017modeling}, used a k-means clustering algorithm to group activities according to trip attributes to synthesize activity chains.
	
	In this paper, we have proposed an algorithm for creating an \gls{abm} for the Greater Melbourne area using a combination of machine learning, probabilistic and gravity-based approaches.
	We combine these techniques in a hybrid model with three primary innovations: 1. when assigning activity patterns, we dispense with the typical nlogit-derived tour-based approach and instead use an activity scheduling model to generate cohort-based individual activity chains for every agent where cohorts are selected based on the activity pattern probabilities; 2. when selecting destinations, we aim to strike a balance between the distance-decay of trip lengths and the activity-based attraction of destination locations; and 3. we take into account the number of trips remaining for an agent so as to ensure they do not select a destination that would be unreasonable to return home from. 
	In this way, our method does not rely on the accuracy and replication of individual travel survey participants and is therefore free from the issues that can arise from replicating the trips of individuals that are outliers. Instead, travel survey and demographic attribute data are used to generate new agents that are demographically representative of the original survey participants and the trips they take.
	Additionally, by selecting destinations in a way that considers trip length and destination location, our model provides greater spatial context to agents' behavior.
	
	In addressing these issues, this research aims to develop an open-source process that generates a synthetic population of agents and an \gls{abm} that is compatible for use with commonly used agent-based modelling software such as MATSim.
	To do this, we use publicly available data from metropolitan Melbourne, Australia.
	Briefly, our process creates a synthetic population of agents with demographic attributes and activity chains derived from publicly available data from the \gls{vista} and from the \gls{abs} Census data drawing on location and mode attributes. 
	
	In the next section, we present the methods.
	We begin by detailing the activity chains undertaken by individuals using trip table data for weekday travel behavior from the VISTA survey in Section \ref{sec:1-setup}.
	Section \ref{sec:2-sample} develops a representative sample of demographic attributes based on ABS census data.
	Section \ref{sec:3-plan} generates trips based on the VISTA data matching activity chains to their time distributions.
	Section \ref{sec:4-match} matches the VISTA activity chains to Census-chains, and Sections \ref{sec:5-locate} and \ref{sec:6-place} details how location and spatial information is assigned to the agents.
	Section \ref{sec:7-time} assigns timing to the agents.
	Results are presented in Section \ref{sec_results} and Section \ref{sec_discussion} discusses key findings.
	
	\section{Method}\label{sec:method} 
	
	Our method is heavily based on daily trips for the Greater Melbourne population sample in \acrfull{vista} 2012-18 data\footnote{\url{https://transport.vic.gov.au/about/data-and-research/vista}}, which has been made openly available by the Department of Transport. \vista sampled approximately 1\% of the Greater Melbourne population using a multi-stage, variable-proportion, clustered sampling of household addresses within \gls{sa1}\footnote{\gls{sa1} regions cover approximately 400 households, there being 10,289 \gls{sa1} regions within Greater Melbourne as per census 2016.} regions. For this work, we used anonymized data for 174,270 trips, representing 49,453 persons from 21,941 households.
	
\begin{figure}[h]
		\centering
		\includegraphics[width=\textwidth]{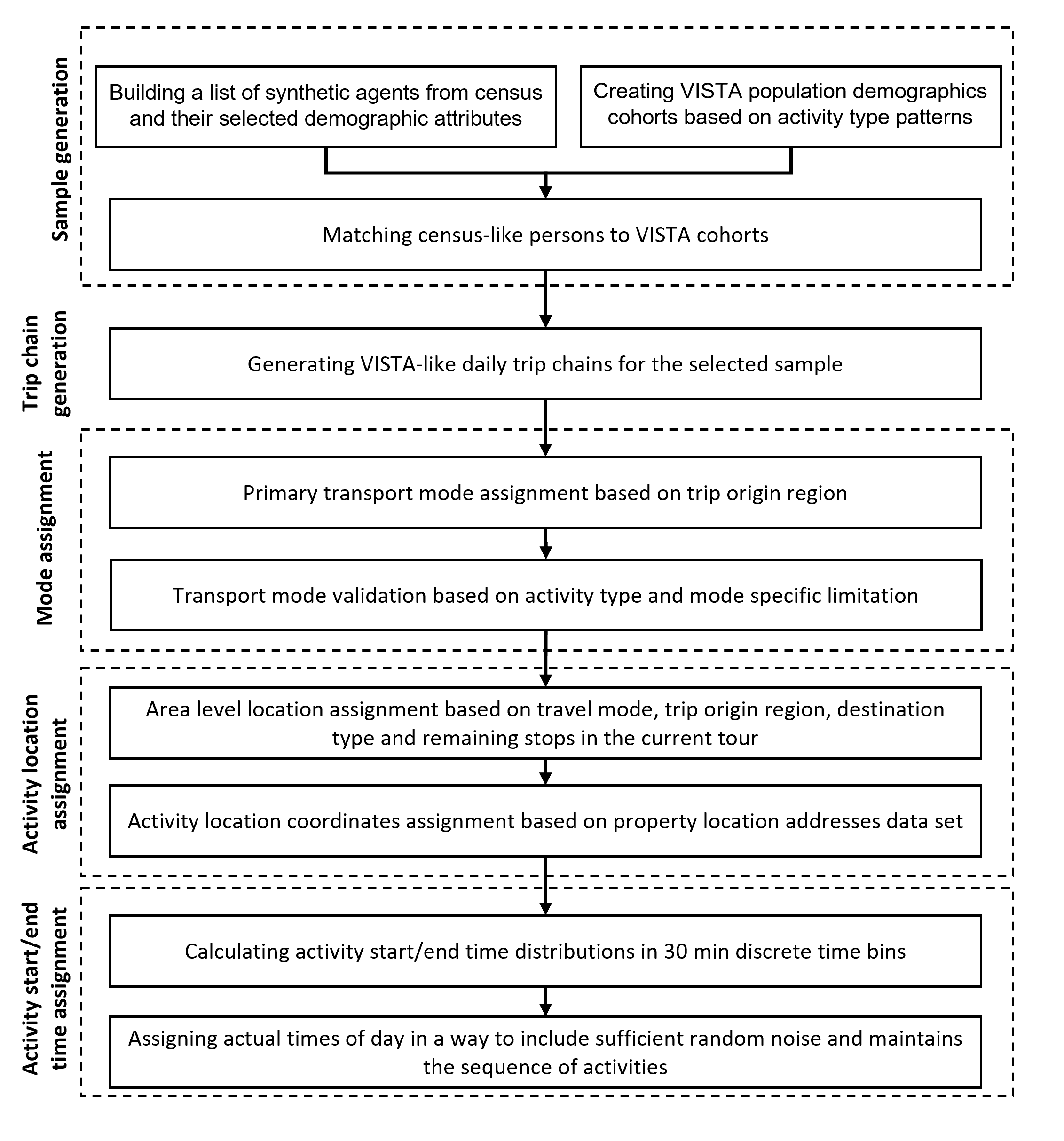}
		\caption{Overview of the activity-based transport demand generation method (Section~\ref{sec:method}).}
		\label{fig:method}
	\end{figure}

	The overall process for building the activity-based transport demand, including the different data used in different steps, is outlined in Figure~\ref{fig:method}. Conceptually, the process takes the daily trips for the Greater Melbourne population sample in \vista, creates an activity generation model from them, and then uses that model to generate as many \vista-like daily travel plans as desired by the user. These activity-trip chains are then matched by demographic attributes to census-like persons sampled from a full synthetic census population for Greater Melbourne made available by \cite{wickramasinghe2020building}. The result is a synthetic population of individuals that have: 
	\begin{inparaenum}[(a)]
		\item demographic attributes consistent with census 2016 at the \gls{sa2} level\footnote{\gls{sa2} approximately covers a typical population center in Australia, there being 307 such in Greater Melbourne according to census 2016.};
		\item sensible home locations assigned from suitable street addresses in those areas; 
		\item daily activity plans consistent with the kinds of daily plans persons of their demographic makeup have in the \vista data; and
		\item activities' times, durations, and locations, as well as mode of travel used to travel between them, consistent with \vista residents of those areas.
	\end{inparaenum}
	
	%\begin{figure}[H]
	
	The process is fully automated and has been optimized to parallelize aspects using multiple cores where available for speedup, making it suitable for use on desktop machines and high performance infrastructure alike.
	To give the reader an appreciation for computing requirements and run times, building a 10\% population sample for Greater Melbourne corresponding to 311,788 unique synthetic agents with their full travel diaries took 
	16 hours and 23 minutes
	on our test machine with 16 cores (Intel i7-6900K 3.20GHz) and 32GB RAM. 
	
	The output of the algorithm is a \gls{csv} file containing the generated daily travel plans of the synthetic travelers, in the form of travel diaries, not too dissimilar to the travel diaries provided in \vista. An example of a generated diary is given in Table~\ref{tab:diary}. The steps of the algorithm incrementally add data to the relevant columns of the table, for all agents. Column headings and the process detailing how the table is populated are explained in subsequent sections. 
	
	\begin{table}[h]
		\centering
		\caption{Example synthetic output travel diary data generated by the algorithm presented in Section~\ref{sec:method}. Note that data consists of 13 columns for the four activities of the agent compressed to fit the page}
		\label{tab:diary}
		\zz{\kk{PlanId}}{161\\161\\161\\161\\}
		\zz{\kk{Activity}}{Home\\Study\\Work\\Home}
		\zz{\kk{StartBin}}{1\\17\\34\\45\\}
		\zz{\kk{EndBin}}{17\\34\\45\\48\\}
		\zz{\kk{AgentId}}{213021342P2848193\\213021342P2848193\\213021342P2848193\\213021342P2848193\\}
		\zz{\kk{SA1}}{21302134204\\21302134242\\21302134115\\21302134204\\}
		\zz{\kk{LocationType}}{home\\education\\work\\home\\}
		\zz{\kk{Mode}}{NA\\car\\car\\car\\}
		\zz{\kk{Distance}}{1519\\2509\\4028\\NA\\}
		\zz{\kk{X}}{304992\\305720\\307321\\304992\\}
		\zz{\kk{Y}}{5804707\\5805729\\5806231\\5804707\\}
		\zz{\kk{StartTime}}{00:03:30\\08:29:00\\17:00:00\\22:07:30\\}
		\zz{\kk{EndTime}}{08:07:00\\16:54:00\\22:02:00\\23:50:00\\}
	\end{table}

	We now describe in detail the different parts of the process of Figure~\ref{fig:method}, which produces an output like that in Table~\ref{tab:diary}.
	
	\subsection{Processing the \vista Trip Table}\label{sec:1-setup}
	
	The first step in building the activity-based transport demand for Greater Melbourne was to process the raw anonymized \vista 2012-18 data. These data are provided in several \acrfull{csv} files and for this study we used what is known in the \vista dataset as the Trip Table (provided in \kk{T\_VISTA1218\_V1.csv}).

	\subsubsection{Understanding \vista Trip Table data}
	
	Table~\ref{tab:vista-trips-sample} shows a sample of the Trip Table data for a household (id: \kk{Y12H0000104}) of three members (ids: \kk{Y12H0000104P01}, \kk{Y12H0000104P02}, \kk{Y12H0000104P03}). 
	Here \kk{PERSID} is the person ID number, \kk{ORIGPURP1} is Origin Purpose (Summary), \kk{DESTPURP1} is Purpose at End of Trip (Summary), \kk{STARTIME} is Time of Starting Trip (in minutes, from midnight), \kk{ARRTIME} is Time of Ending Trip (in minutes, from midnight), and \kk{WDTRIPWGT} is Trip weight for an `Average weekday' of the combined 2012-18 dataset, using the Australian Standard Geographical Classification (ASGC)\footnote{\url{https://www.abs.gov.au/websitedbs/D3310114.nsf/home/Australian+Standard+Geographical+Classification+(ASGC)}}. 
	We focus the following discussion on an average weekday, but the technique may be applied in the same way for the `Average weekend day' data rows, given by the corresponding \kk{WEJTEWGT} column.
	
	\begin{table}[h]
		\centering
		\caption{Trips of an example household \kk{Y12H0000104} from the \vista Trip Table}
		\label{tab:vista-trips-sample}
		{\footnotesize
			\begin{tabular}{|llllll|}
				\hline
				\kk{PERSID} & \kk{ORIGPURP1} & \kk{DESTPURP1} & \kk{STARTIME} & \kk{ARRTIME} & \kk{WDTRIPWGT}\\
				\hline\hline
				Y12H0000104P01& At Home      & Work Related & 420     & 485      & 83.77\\
				Y12H0000104P01& Work Related & Go Home      & 990     & 1065     & 83.77\\
				Y12H0000104P02& At Home      & Work Related & 540     & 555      & 86.51\\
				Y12H0000104P02& Work Related & Buy Something& 558     & 565      & 86.51\\
				Y12H0000104P02& Buy Something& Go Home      & 570     & 575      & 86.51\\
				Y12H0000104P02& At Home      & Buy Something& 900     & 905      & 86.51\\
				Y12H0000104P02& Buy Something& Go Home      & 910     & 915      & 86.51\\
				Y12H0000104P03& At Home      & Work Related & 450     & 480      & 131.96\\
				Y12H0000104P03& Work Related & Go Home      & 990     & 1020     & 131.96\\
				\hline
			\end{tabular}
		}
	\end{table}
	
	Table~\ref{tab:vista-trips-sample} shows the complete set of Trip Table attributes that our algorithm uses to generate \vista-like activity/trip chains. 
	Notably, this stage disregards all geospatial information from the records, focusing only on the sequence of activities and trips. 
	This is because the intent is to initially generate location-agnostic \vista-like activity/trip chains, and then in subsequent steps of the algorithm, place these activities and trips in the context of the geographical home location assigned to the synthetic agent. 
	Also note that no information about the mode of transportation is retained at this stage. 
	Again, this will be introduced in the context of the home location later in the process.
	
	\begin{figure}[h]
		\centering
		\input{figs/vista-trip}
		\caption{Ordered sequence of activities (circles) and trips (arrows) for anonymous person \kk{Y12H0000104P02} in the \vista Trip Table}
		\label{fig:vista-person-trips}
	\end{figure}

	Figure~\ref{fig:vista-person-trips} gives a visual representation of the sequence of activities and trips of the example person \kk{Y12H0000104P02} of Table~\ref{tab:vista-trips-sample} with each row of the table represented by a numbered arrow in the figure. 
	Person \kk{Y12H0000104P02}'s day could be summarized as: left home at 9am (540 minutes past midnight) and 15 minutes later performed a quick work-related activity that lasted three minutes (maybe to the local post office?); went back home via a quick seven-minute stop to buy something (a morning coffee perhaps?); stayed at home from 9:35am (575 minutes past midnight) to 3pm (900 minutes past midnight); did another quick 15 minute round trip to the shops; then stayed home for the rest of the day.
	
	The trips of sample person \kk{Y12H0000104P02} highlight some important choices that must be made when interpreting the data. 
	For instance, did the person go to different shops (as we suggest in Figure~\ref{fig:vista-person-trips}) or the same one? 
	Does the same assumption apply for all kinds of trips? In general, we applied the following rules to multiple trips for the same kind of activity during the day.
	
	\begin{itemize}
		\item All trips that start and end at a home-related activity (\kk{ORIGPURP1} or \kk{DESTPURP1} contain the string `Home') are assumed to be associated with the same home location.
		\item All sub-tours (sequences of activities starting and ending at home, such as the morning and afternoon sub-tours of person \kk{Y12H0000104P02}) that contain multiple work-related trips are assumed to be associated with a single work location, however work locations may change between two sub-tours (i.e., whenever the agent arrives at home).
		\item All other trips (including shopping-related trips) are assumed to be potentially associated with different locations, even if performed within the same sub-tour.
	\end{itemize}

	%-------------------------------------------------------------------------------------
	\subsubsection{Extracting daily activities from the Trip Table}\label{sec:activity-table}
	
	Each row in Table~\ref{tab:vista-trips-sample} gives the start and end time of a single trip, and it is easy to see that the difference between the start time of one trip and the end time of the \textit{preceding} trip of the person is, in fact, the duration of the activity between those two trips. 
	The first trip in the chain does not have a preceding activity, of course, but here the concluding activity can safely be assumed to have a start time of midnight. 
	Therefore, this knowledge can be used to transform the Trip Table into a corresponding \textit{activity-table}.
	Table~\ref{tab:vista-activities-sample} shows this transformation for the trips of our sample household.
	
	\begin{table}[h]
		\centering
		\caption{Activities of example household \kk{Y12H0000104} derived from Trip Table}
		\label{tab:vista-activities-sample}
		{\footnotesize
			\begin{tabular}{|lllll|}
				\hline
				\kk{PERSID} & \kk{ACTIVITY} & \kk{START.TIME} & \kk{END.TIME}  & \kk{WDTRIPWGT}\\
				\hline\hline
				Y12H0000104P01 & At Home & 0 & 420 & 83.77\\
				Y12H0000104P01 & Work Related & 485 & 990 & 83.77\\
				Y12H0000104P01 & Go Home & 1065 & 1439 & 83.77\\
				Y12H0000104P02 & At Home & 0 & 540 & 86.51\\
				Y12H0000104P02 & Work Related & 555 & 558 & 86.51\\
				Y12H0000104P02 & Buy Something & 565 & 570 & 86.51\\
				Y12H0000104P02 & At Home & 575 & 900 & 86.51\\
				Y12H0000104P02 & Buy Something & 905 & 910 & 86.51\\
				Y12H0000104P02 & Go Home & 915 & 1439 & 86.51\\
				Y12H0000104P03 & At Home & 0 & 450 & 131.96\\
				Y12H0000104P03 & Work Related & 480 & 990 & 131.96\\
				Y12H0000104P03 & Go Home & 1020 & 1439 & 131.96\\
				\hline
			\end{tabular}
		}
	\end{table}
	
	The activities table produced from the entire Trip Table then provides the raw input that forms the basis for our activity-based plan generation algorithm.
	
	%-------------------------------------------------------------------------------------
	\subsubsection{Simplifying activity labels}
	
	The number of unique activity labels present in the activity-table derived in the previous step was reduced by grouping related labels into simpler tags.
	We also renamed some labels for clarity.
	Table~\ref{tab:vista-labels-replacement} shows the specific text replacements that we performed in the \kk{ACTIVITY} column of our activity-table from the previous step.
	The resulting activity-table has \A unique activity types, being: Home, (Transport) Mode Change, Other, Personal, Pickup/Dropoff/Deliver, Shop, Social/Recreational, Study, With Someone, Work.
	
	\begin{table}[h]
		\centering
		\caption{Label simplification performed on the Trip Table derived activity names}
		\label{tab:vista-labels-replacement}
		{\footnotesize
			\begin{tabular}{|l|l|}
				\hline
				\textbf{Original Trip Table label} & \textbf{Replacement label}\\
				\hline\hline
				At Home ; Go Home ; Unknown Purpose (at start of day) & Home\\
				Social ; Recreational & Social/Recreational\\
				Pick-up or Drop-off Someone ; Pick-up or Deliver Something & Pickup/Dropoff/Deliver\\
				Other Purpose ; Not Stated & Other\\
				Personal Business & Personal\\
				Work Related & Work\\
				Education & Study\\
				Buy Something & Shop\\
				Change Mode & Mode Change\\
				Accompany Someone & With Someone\\
				\hline
			\end{tabular}
		}
	\end{table}
	
	%-------------------------------------------------------------------------------------
	\subsubsection{Calculating activity start/end time distributions in \T discrete time bins}
	
	The final step in the \vista data processing is to calculate the start and end time distributions for each activity type throughout the day--where the day is split into \T discrete time bins of equal size.
	The parameter \T gives a way of easily configuring the desired precision of the daily plan generation step of the algorithm (Section~\ref{sec:3-plan}).
	Higher values of \T allow the algorithm to seek higher precision in the generated activity start/end times but can lead to more variance in error, while lower values seek coarser precision, which is easier to achieve, resulting in lower error.
	What value gives a good balance between precision and error can be determined from experimentation.
	In this work, we used \T$=48$, i.e., we break up the day into 48 time bins of 30 minutes each.
	
	\begin{figure}[h]
		\centering
		\input{figs/matrix}
		\caption{Structure of matrix \D for storing start(end) time distributions for $\vert\mathcal{A}\vert$ activities against \T time bins of the day}
		\label{fig:matrix}
	\end{figure}

	Calculation of the activities' start(end) time distribution is done by first counting, for each activity, the number of instances of activity start(end) in every time bin.
	To do so, a matrix \D (being \Ds for start time distributions and \De for end time distributions) is created with \A rows of unique activity types and \T columns for every time bin (Figure~\ref{fig:matrix}). Then for every row $r$ in the activity-table, the value of the corresponding cell in \D is updated, given by the \D-row that matches the \kk{ACTIVITY} label in $r$ and the \D-column corresponding to the time bin for the start(end) time of the activity in $r$.
	The value of this determined cell is then incremented by the value of the \kk{WDTRIPWGT} column in $r$, which gives the frequency of this activity in the full population (remembering that the \vista Trip Table represents a 1\% sample).
	This output is then stored in \csv format, to be used by subsequent steps of the algorithm.
	Note that the saved end-time distribution table \De has \A$\times$\T rows, since the end time is saved for every activity type (Table~\ref{tab:vista-labels-replacement}) and every start time bin. In other words, for end times, we store \T matrices of the type shown in Figure~\ref{fig:matrix}.
	
\begin{figure}[h]
		\centering
		\includegraphics[width=\textwidth]{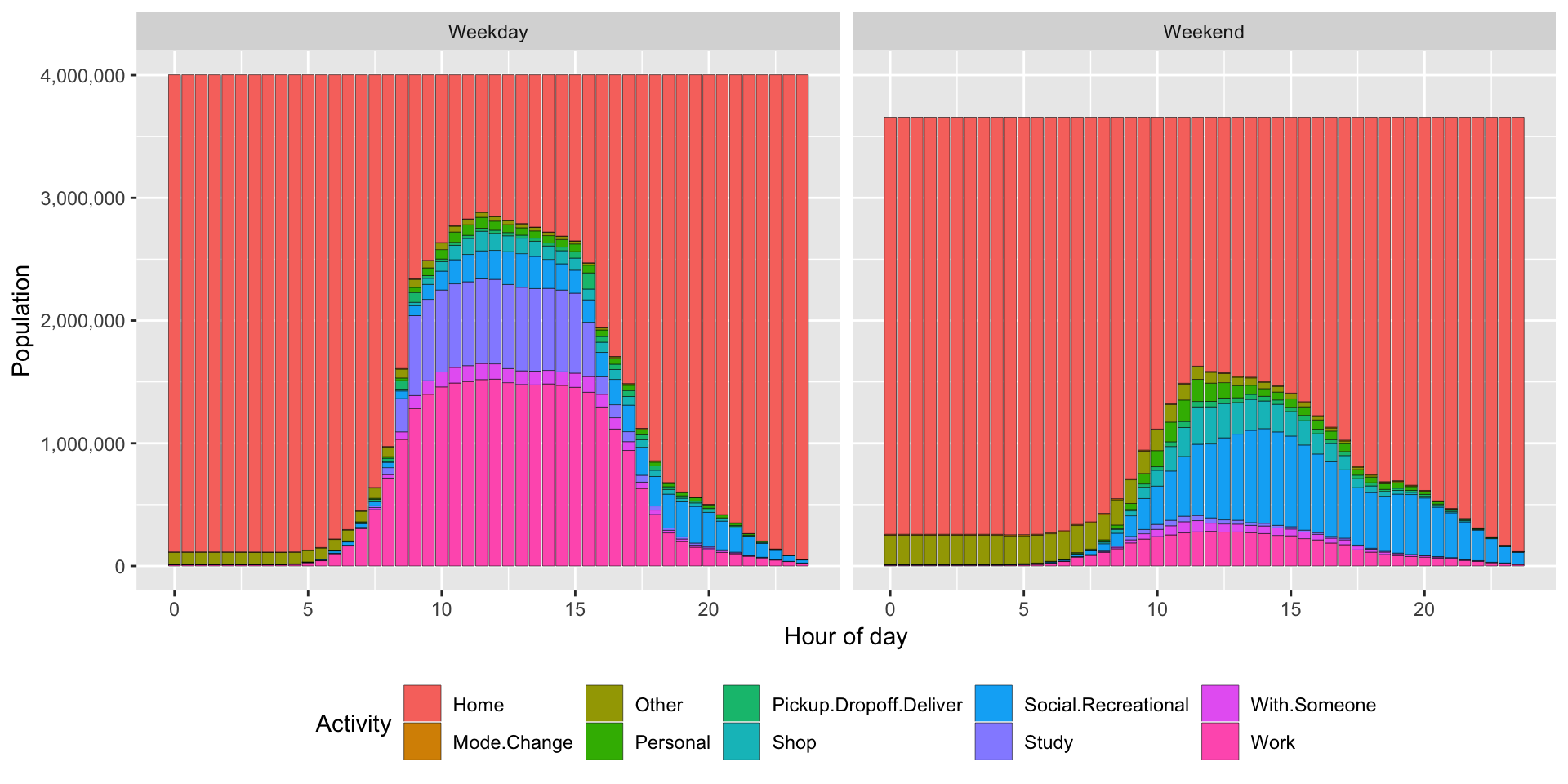}
		\caption{Simplified \vista Trip Table derived activities in \T$=48$ discrete time bins}
		\label{fig:vista-activity-time-bins}
	\end{figure}

	Figure~\ref{fig:vista-activity-time-bins} shows a consolidated view of the simplified activities of the Melbourne population across the day split into \T$=48$ discrete time bin distributions, computed separately for the weekday and weekend rows of the activity-table we derived from Trip Table. Each split bar shows the proportion of the population performing the different activities in \A during the corresponding time bin. 
	
	\subsubsection{Generating \vista population cohorts}\label{sec:cohorts}
	
	The process explained so far in Section~\ref{sec:1-setup} shows how the \vista Trip Table, bar weekend trips, is used to compute distributions of activities in discrete time bins of the day. We now describe a slight modification to the process, to account for differences in activity profiles across population subgroups, or \textit{cohorts}. For instance, the likelihood of work activities in children and older adults is lower than working-age adults. By identifying cohorts, and creating separate activity generation models for each based on their activity likelihoods, we get better accuracy overall for the population than can be achieved by a single model.  
	
	It is well understood that the kinds of activities people do can be shaped by individual, social, and environmental factors~\citep{bautista2020urban,grue2020exploring}.
	Not all of these will be known and/or related variables available in the \vista data. Nevertheless, it is clear that some steps can be taken to classify observed behaviors into groups given the variables we do have. 
	
\begin{figure}[h]
		\centering
		\includegraphics[width=\textwidth]{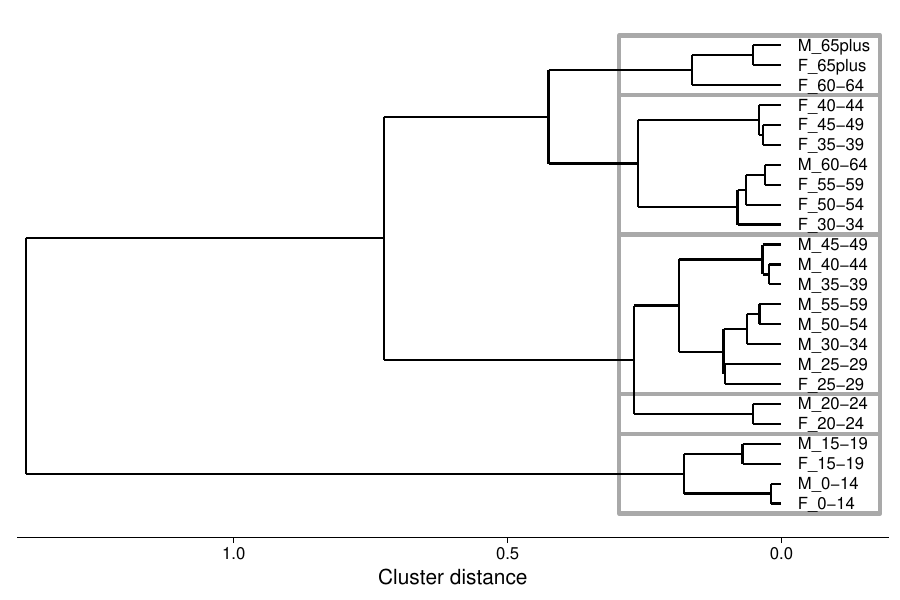}
		\caption{Dendrogram of hierarchical cluster analysis. Grey rectangles indicate cluster membership by age and sex based on probabilities of engaging in Work, Study, Shop, Personal and Social/Recreational activities.}
		\label{fig:dendrogram}
	\end{figure}

	We implemented a simple classification based on the demographic attributes of sex and age, to find distinct groupings that exhibit similar activity profiles. Specifically, participants were broken into five-year age groups, with the exception of groups 0--14 and 65 and over. Probabilities of engaging in the activities Work, Study, Shop, Personal, and Social/Recreational were then calculated for each of these 24 groups based on weekday trips. Hierarchical clustering was then applied to the dataset using Ward's method, producing the dendrogram shown in Figure \ref{fig:dendrogram}, with the main output being cluster membership indicating similar activity probability profiles by age and sex.
	
	The gap statistic of the clustering process indicated that five unique cohorts would be optimal. Each cohort identified participants with similar activity patterns based on age and sex as the output. Our final step then was to filter the \vista Trip Table records on those attributes and store the result as partial tables, one table per cohort. The process described so far in Section~\ref{sec:1-setup} was therefore applied to each partial trip table separately rather than the \vista Trip Table directly--as suggested earlier to keep the explanation simple--giving activity distribution tables by time of day per cohort. 
	
	\subsection{Creating a representative sample of census-like persons}\label{sec:2-sample}
	
	This step of the algorithm is concerned with allocating the right kinds of persons to the right statistical areas in Greater Melbourne as per the Australian Census 2016.
	The output of this step is a synthetic population consisting of a list of uniquely identified agents each described by a valid street address representing their home location, and demographic attributes of age and gender that, when aggregated, match the demographic distributions reported in the census at the \gls{sa2} level. 
	
	A full synthetic population for Greater Melbourne based on ABS Census 2016 was created by \cite{wickramasinghe2020building}. Their output population is made available through their GitHub code repository\footnote{\url{https://github.com/agentsoz/synthetic-population}}. It consists of a relational database of unique persons in unique households assigned to known street addresses. For convenience, their database is supplied in two CSV files containing `persons' and `households' respectively, separately for each of the 307 \gls{sa2} areas in Greater Melbourne. 
	
	We obtain our census-based synthetic population by sampling a desired number of agents from \cite{wickramasinghe2020building}’s population files. For instance, to build a 10\% sample of the Greater Melbourne population, we randomly sample 10\% from each of the 307 \gls{sa2} level `persons' files. 
	This gave us our base census-like individuals at home locations in Greater Melbourne.
	Subsequent steps assign \vista-like trips to these agents, representing the kinds of trips agents of their demographic makeup in the given \gls{sa2} undertake in their day as per \vista data. 
	
	\subsection{Matching census-like persons to VISTA cohorts}\label{sec:4-match}
	
	In this step of the algorithm, we assign one of the \vista groups (cohorts) previously calculated (Section~\ref{sec:cohorts}), to the census-like people from the last step (Section~\ref{sec:2-sample}) based on their demographic profile. This is required so that in the subsequent step we can apply group-appropriate trip generation models for assigning representative trips and activities to those individuals.
	
	The \textit{matching} of persons to cohorts itself is a straightforward process. Since each cohort is fully defined by a gender and age range, then the cohort of a person can be determined by a simple lookup table. The output of this step is the addition of a new attribute to each person, indicating the cohort to which they belong. 
	
	In the final output of the algorithm, as shown in Table~\ref{tab:diary}, this step is responsible for populating column \kk{AgentId}, that gives a unique identifier for a synthetic census-like traveler sampled in the earlier step (Section~\ref{sec:2-sample}), and was matched to the \vista-like activity chain given by column \kk{PlanId}.
	
	\subsection{Generating VISTA-like daily trips}\label{sec:3-plan}
	
	The algorithm described here does not consider population groups, instead the process being described is applied separately to the cohorts identified in Section~\ref{sec:cohorts}. Additionally, the algorithm generates activity chains based on activity schedules, and therefore does not assign activities based on tours.
	
	Algorithm~\ref{alg:plan-generation} gives the pseudo-code of our algorithm for generating VISTA-like activity chains. The objective of the algorithm is to generate a sequence of \N activity chains in such a way that, when taken together, the list \C of generated activity chains achieves the target activity start(end) time distributions given by the \D-matrices \Ds and \De. Here, the rows of the \D-matrices give the  activities and the columns give their distribution over the time bins of the day. 
	
	The general intuition behind the algorithm is to repeatedly revise the desired distributions by taking the difference $\Delta$ between the presently achieved distributions (\Dso and \Deo) and the target distributions (\Ds and \De) so that over-represented activities in a given time bin are less likely to be generated in subsequent iterations while under-represented activities become more likely. This allows for dynamic on-the-fly revision where the algorithm is continuously looking to correct towards the moving target distributions with every new activity chain it generates. This approach works well in adapting output to the target \D-matrices, and the generation error decreases asymptotically as the number \N of generated activity chains increases. 
	
	\input{figs/algo-step3}

	We describe in Algorithm~\ref{alg:plan-generation} the steps for matching to the target start time distribution matrix \Ds and note that the steps are the same for matching to the end time distribution matrix \De.
	The process starts by initializing an empty list \C for storing the activity chains to generate, a corresponding empty matrix \Dso for recording the start-time distributions of the activities in \C, and another empty matrix $\Delta$ for storing the difference from the target distributions in \Ds (lines 1--3). Both \Dso and $\Delta$ have the same dimensions as \Ds as shown in Figure~\ref{fig:matrix}. The following steps (lines 4--34) are then repeated once per activity chain, to generate \N activity chains. 
	
	Prior to generating a new activity chain, the difference matrix $\Delta$ is updated to reflect the current deviation from the desired distribution \Ds (lines 5--9), ensuring that zero-value cells in \Ds are also zero-value in $\Delta$, and normalizing the  row vectors to lie in the range [0,1]. This means that any zero-value cells in $\Delta$ are either so because the activity (row) does not occur in that time bin (column), or the proportion of the given activity in the given time bin for the population generated thus far either perfectly matches the desired or is over-represented. On the other hand, values tending to $1.0$ indicate increasing levels of under-representation. 
	
	The algorithm then traverses the time bins sequentially from beginning of the day to end (line 10, 12) as follows. 
	We first check if the proportion of activities (across all activities) that start in the current time bin, is already at or above the desired level, and if found to be so we skip to the next time bin (lines 13--17). 
	Otherwise, we extract from $\Delta$, a vector $\delta$, to give the difference from desired for all activities, for the current step (line 18). 
	If $\delta$ is zero because it is also zero in the desired distribution \Ds then we skip to the next time bin (lines 19--20), else we use $\delta$ to probabilistically select a corresponding activity and set its start time to the current time bin (lines 22--23). 
	This procedure results in a higher chance of the selection of under-represented activities. The end time for the selected activity is chosen probabilistically from the probabilities of the given activity ending in the remaining time bins of the day, when starting in the current time bin (lines 24--26).
	The generated activity with its allocated start and end bins is added to the trip chain $\Phi$ (line 27), and we skip to the activity end time bin (line 28) to continue building the chain.
	
	The trip chain $\Phi$ thus generated is then compressed by collapsing consecutive blocks of the same activity into a single activity that starts in the time bin of the first occurrence and finishes in the time bin of the last (line 31). If $\Phi$ was empty to begin with, i.e., no activity was generated in the preceding loop, then we just assign a $Home$ activity lasting the whole day (line 30). The trip chain $\Phi$ is then added to our list \C (line 32). Finally, the counts of the generated activities in $\Phi$ are incremented in \Dso, to be used do update the difference matrix $\Delta$ before generating the next activity chain. 
	
	In the final output of the algorithm, an example of which is given in Table~\ref{tab:diary}, this step is responsible for populating columns \kk{PlanId} (a unique identifier for the generated activity chain), \kk{Activity} (the sequence of activities in the chain), \kk{StartBin} (the start time bin for each activity, being an integer between 1 and 48, representing the 30-minute blocks of the day), and \kk{EndBin} (the corresponding end time bins for the activities).
	
	\subsection{Assigning statistical areas (SA1) to activities}\label{sec:5-locate}
	At this point, each agent now has a home \gls{sa1} and a list of activities they conduct throughout the day.
	Candidate locations for these activities were selected from the endpoint nodes of non-highway edges (i.e., edges with a speed of 60km/h or less and accessible by all modes) of the \gls{osm}-derived transport network generated by \cite{jafari_building_2022}.
	It is important to note that these non-highway edges were already densified, with additional nodes added every 500 meters.
	These locations were selected to ensure that any activity location would be reachable from the network without any non-network travel movement (i.e., bushwhacking/teleporting).
	These locations were then classified according to the land use category of the mesh block they are located within.
	To facilitate the matching of location types to activity types, the mesh block categories were simplified into five types: Home, Work, Education, Commercial, and Park (see Table \ref{tab:vista-category-reclass}). 
	%This resulted in 256,311 unique destination locations.
	
	\begin{table}[h]
		\centering
		\caption{VISTA category reclassification}
		\label{tab:vista-category-reclass}
		{\footnotesize
			\begin{tabular}{|p{0.20\linewidth} p{0.26\linewidth} p{0.38\linewidth}|}
				\hline
				\kk{Location categories} & \kk{Meshblock categories} & \kk{\vista activities} \\
				\hline\hline
				Home                & Residential, Other, Primary Production                                  & Home                                                                             \\
				Work                & Commercial, Education, Hospital/Medical, Industrial, Primary Production & Other, Pickup/Dropoff/Deliver, With Someone, Work                                \\
				Education           & Education                                                               & Other, Pickup/Dropoff/Deliver, Study, With Someone                               \\
				Commercial          & Commercial                                                              & Other, Personal, Pickup/Dropoff/Deliver, Shop, Social/Recreational, With Someone \\
				Park                & Parkland, Water                                                         & Other, Pickup/Dropoff/Deliver, Social/Recreational, With Someone                 \\               
				\hline
			\end{tabular}
		}
	\end{table}
	
	Transport mode is then assigned sequentially to each trip, along with \gls{sa1} region to non-home activities. The specifics are detailed in Algorithm \ref{alg:locate}, but broadly, transport mode is selected first. This then allows for region selection based on the likely travel distance for that mode, as well as the relative attractiveness of potential destinations for the chosen trip purpose. 
	
	\input{figs/algo-step5}
	
	\subsubsection{Mode selection}\label{sec_modeSelection}
	Transport mode selection is performed by the getMode function, which selects from the possibilities of walk, bike, PT (Public Transit), or car.
	This function takes the current region as input to ensure that local variation in mode choice is present in the agents' behavior.
	Specifically, some modes, such as walking and public transit are more popular towards the inner city, whereas driving is preferred by residents of the outer suburbs.
	
	The first run of the function for each agent sets their primary mode, which is the initial mode used when an agent leaves home.
	Primary mode is used as an input of the getMode function to ensure vehicle use (i.e., car or bike) is appropriate. Specifically, if a vehicle is not initially used by an agent, then it is not possible to select one at any other point throughout the day.
	However, it is possible for agents to switch from a vehicle to walking or public transit.
	To prevent agents leaving vehicles stranded, they must return to that region so they may use the vehicle to return home.
	Additionally, walking and public transit may be freely switched between, so long as the final trip home utilizes public transit. 
	
	Including the current region in the mode assignment ensures that the likely mode chosen is appropriate to the types of trips undertaken in the underlying travel survey data between origins and destinations.  For example, if in the dataset we observed nearby a coffee shop and visiting it typically involved a walk trip, then walking is likely to be the selected mode. For other less walkable neighbourhoods, where the distance to the coffee shop is further away, the same coffee shop trip may likely be done by car. 
	
	The mode choice probabilities used by the getMode function were generated for each \gls{sa1} region by analyzing the proportions of modes chosen by the participants of the \vista travel survey.
	For this, the full travel survey was used, meaning that origin (\kk{ORIGLONG}, \kk{ORIGLAT}) and destination (\kk{DESTLONG}, \kk{DESTLAT}) coordinates were available.
	Trips were filtered to weekdays (using \kk{TRAVDOW}) within the Greater Melbourne region and recorded with their origin location, survey weight (\kk{CW\_WDTRIPWGT\_SA3}), and transport mode (\kk{LINKMODE}, which was reclassified to match the modes included in our transport demand model).
	Survey weight is a number that indicates how representative each entry is of the Victorian population during a weekday, and was used in calibrating accurate mode choice proportions.
	To determine the mode proportions, Kernel Density Estimation (KDE) was calculated at each candidate location for each transport mode, using a weighted Gaussian kernel with a bandwidth of 750 meters, to be aggregated up to \gls{sa1} and converted to a percentage.
	Destination locations were used instead of creating a density raster as it is much faster to compute, and only produces density calculations at places agents are able to travel.
	
	Selecting an appropriate bandwidth for density calculations is important as smaller bandwidths show greater local variation but have fewer points to use, and larger bandwidths use more points, but show more general trends.
	In this case, 750 meters was chosen based on calibration. Specifically, a variety of potential bandwidths were chosen, with their mode choice percentages aggregated to the 40 \gls{sa3} regions comprising Greater Melbourne.
	These percentages were then compared with values obtained by aggregating the weighted \gls{vista} mode choice proportions to \gls{sa3} regions in order to select the bandwidth with the best fit.
	\gls{sa3} regions were chosen for calibration as that is the statistical granularity at which the \gls{vista} travel survey was weighted.

	\subsubsection{Destination selection}\label{sec_destinationSelection}
	
\begin{figure}[H]
		\centering
		\begin{tabular}{cc}
			\includegraphics[width=0.45\columnwidth]{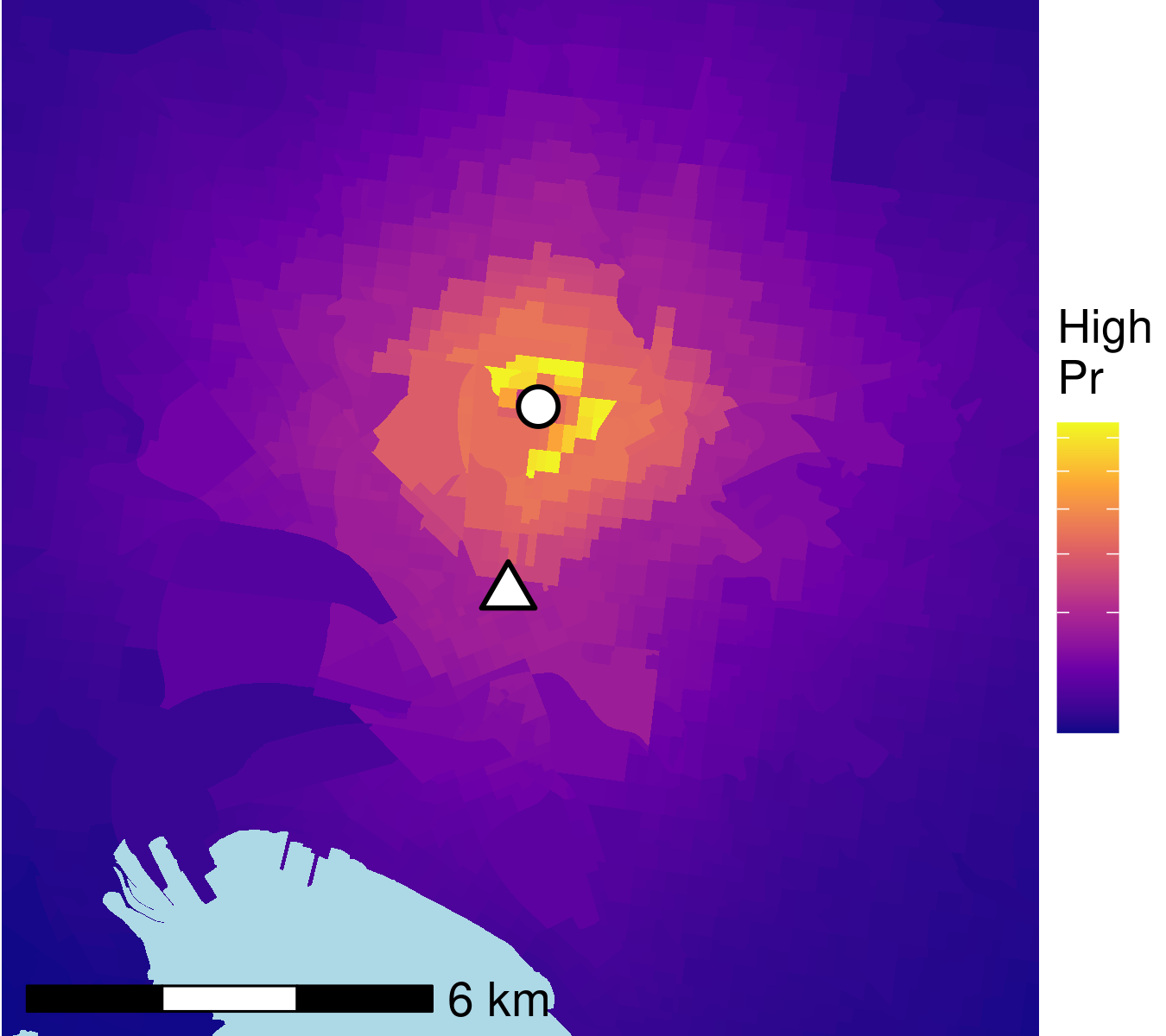} &
			\includegraphics[width=0.45\columnwidth]{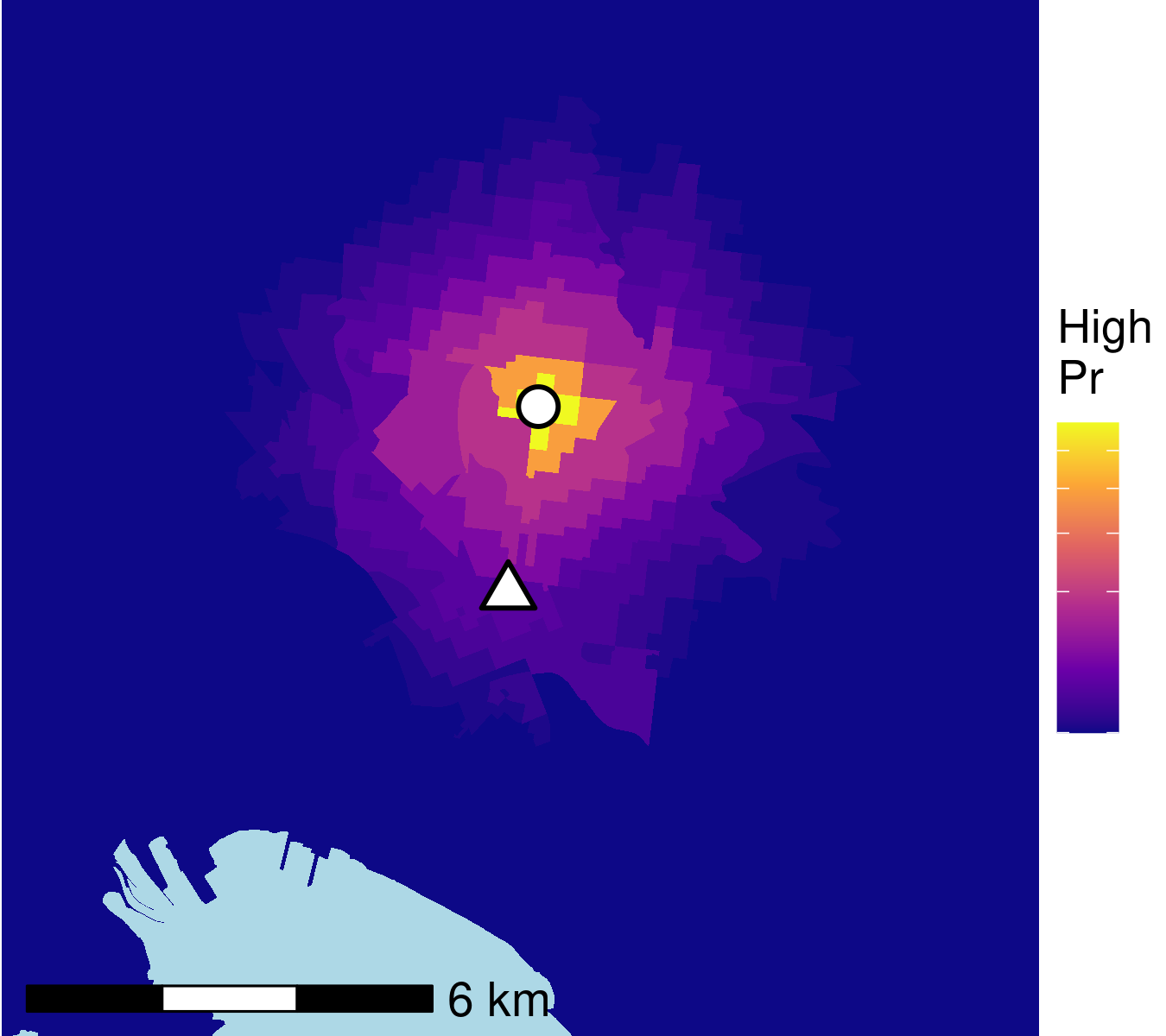}\\
			(a) & (b)\\
			\includegraphics[width=0.45\columnwidth]{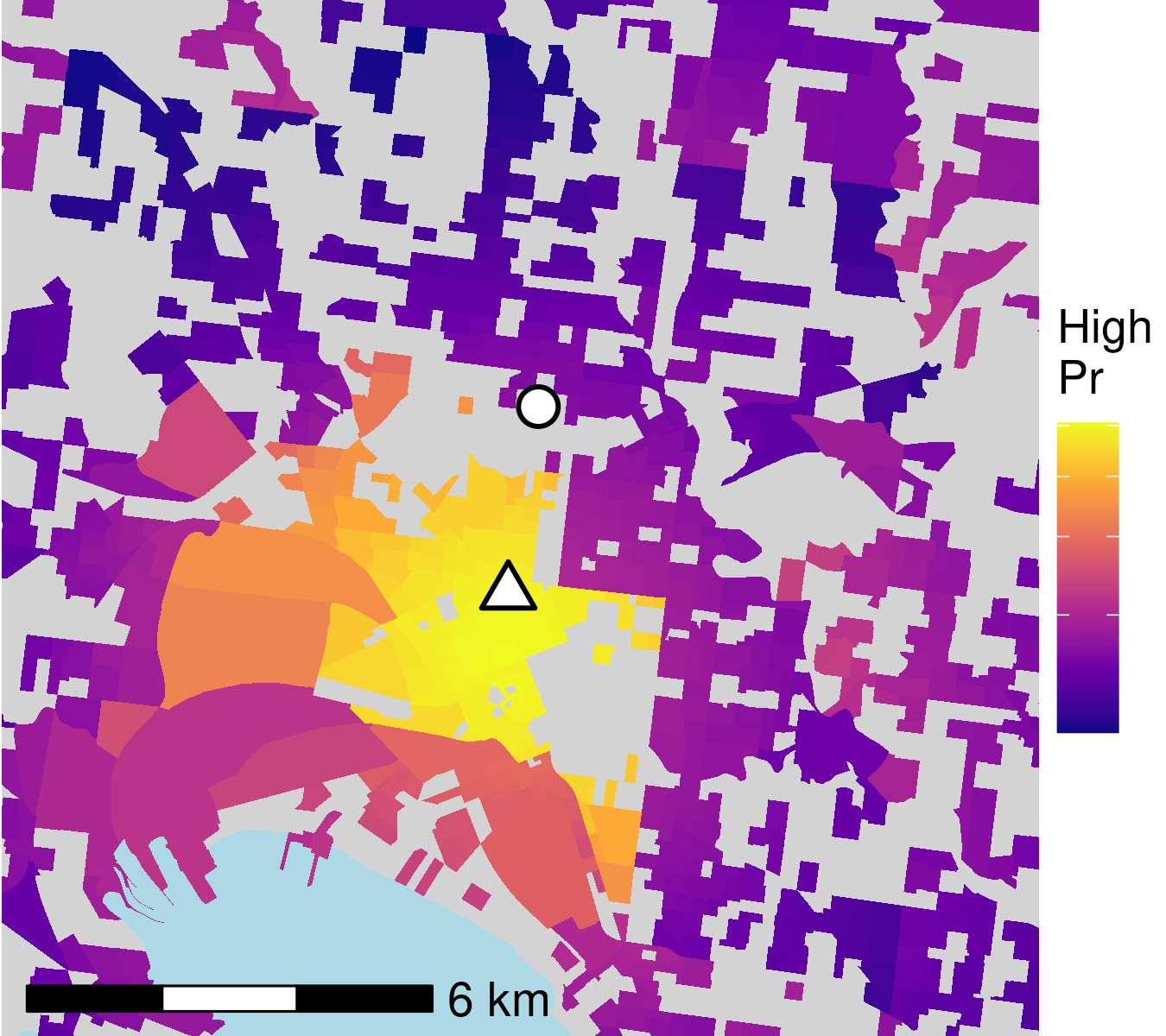} &
			\includegraphics[width=0.45\columnwidth]{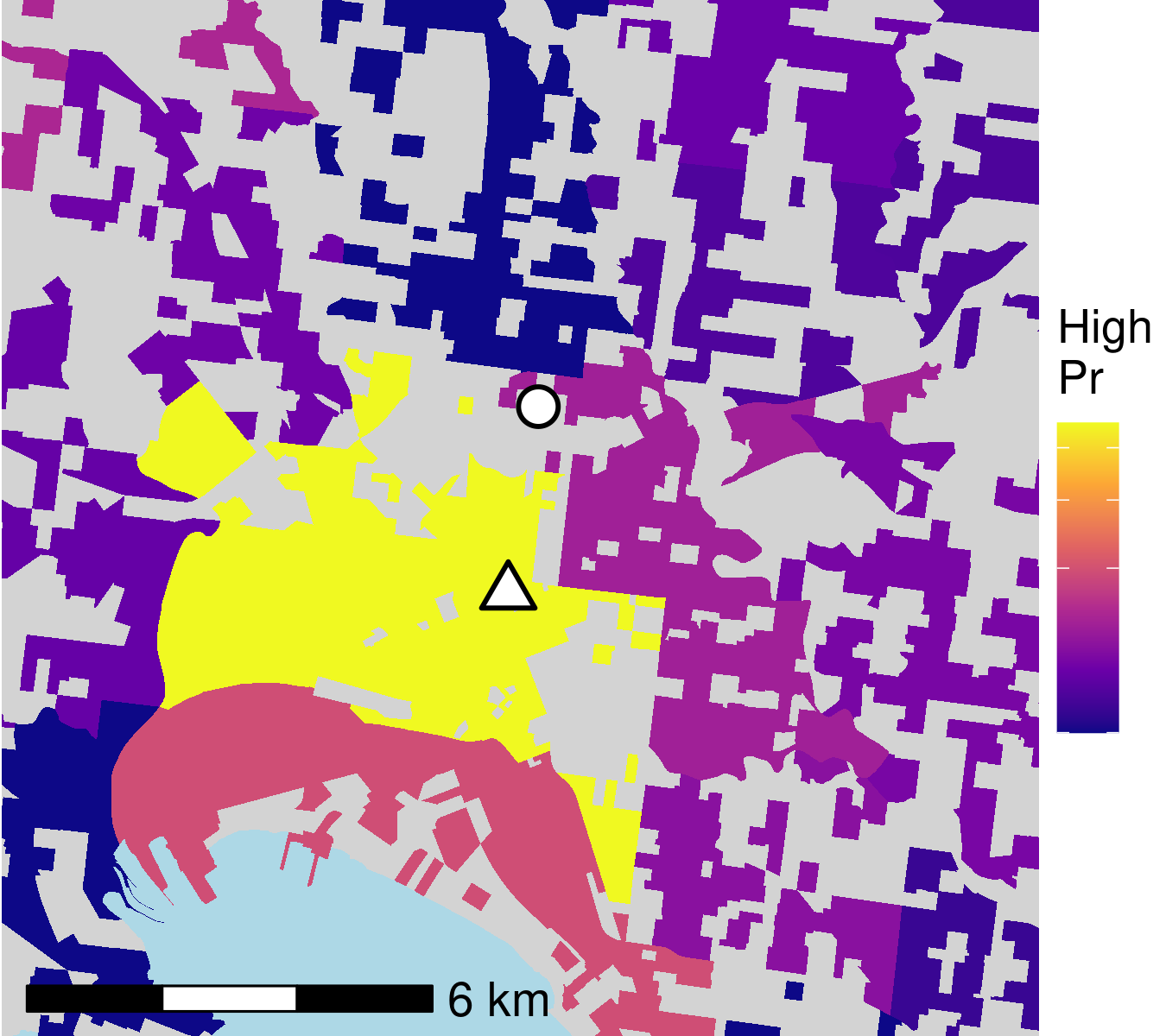}\\
			(c) & (d)\\
			\includegraphics[width=0.45\columnwidth]{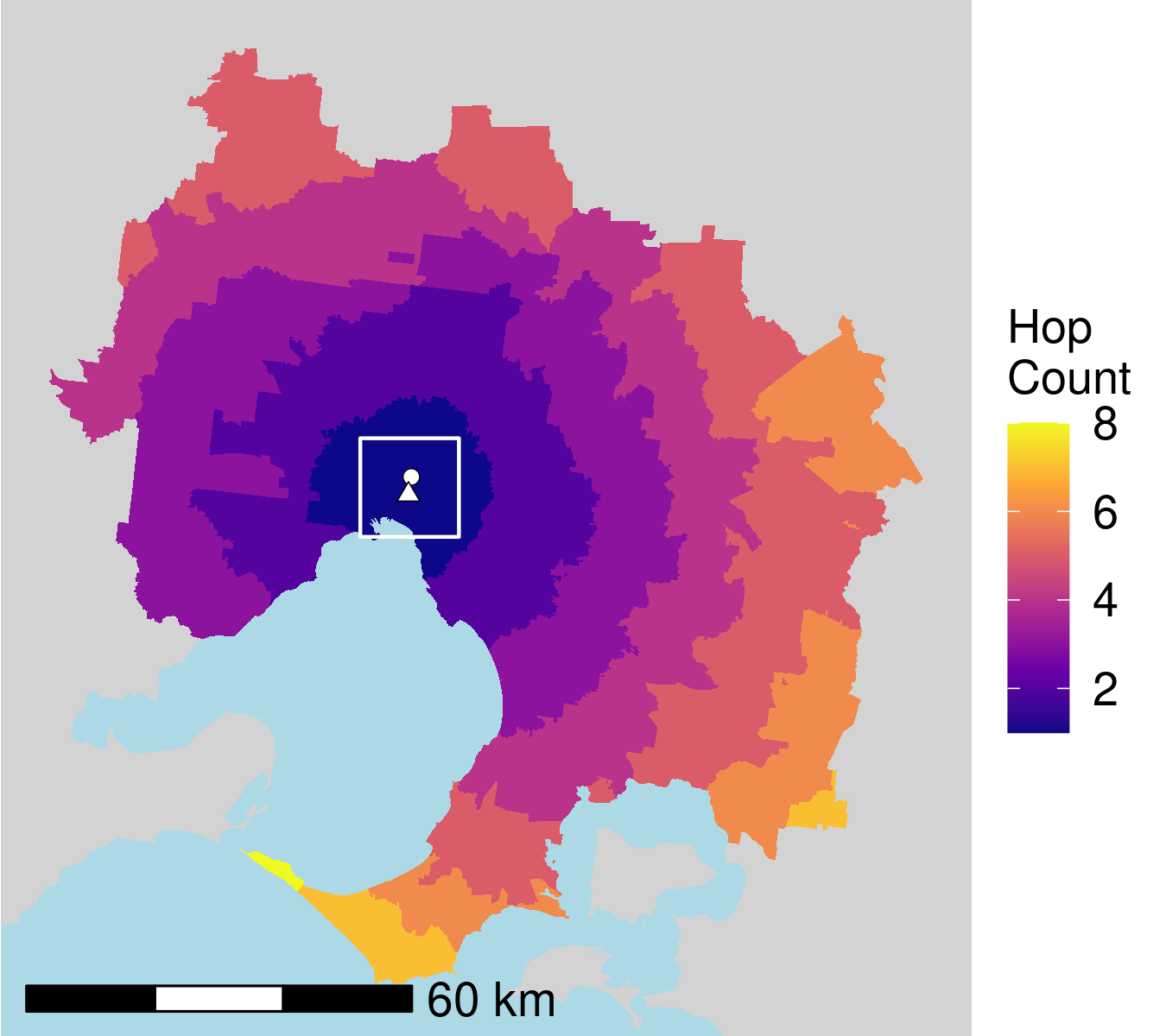} &
			\includegraphics[width=0.45\columnwidth]{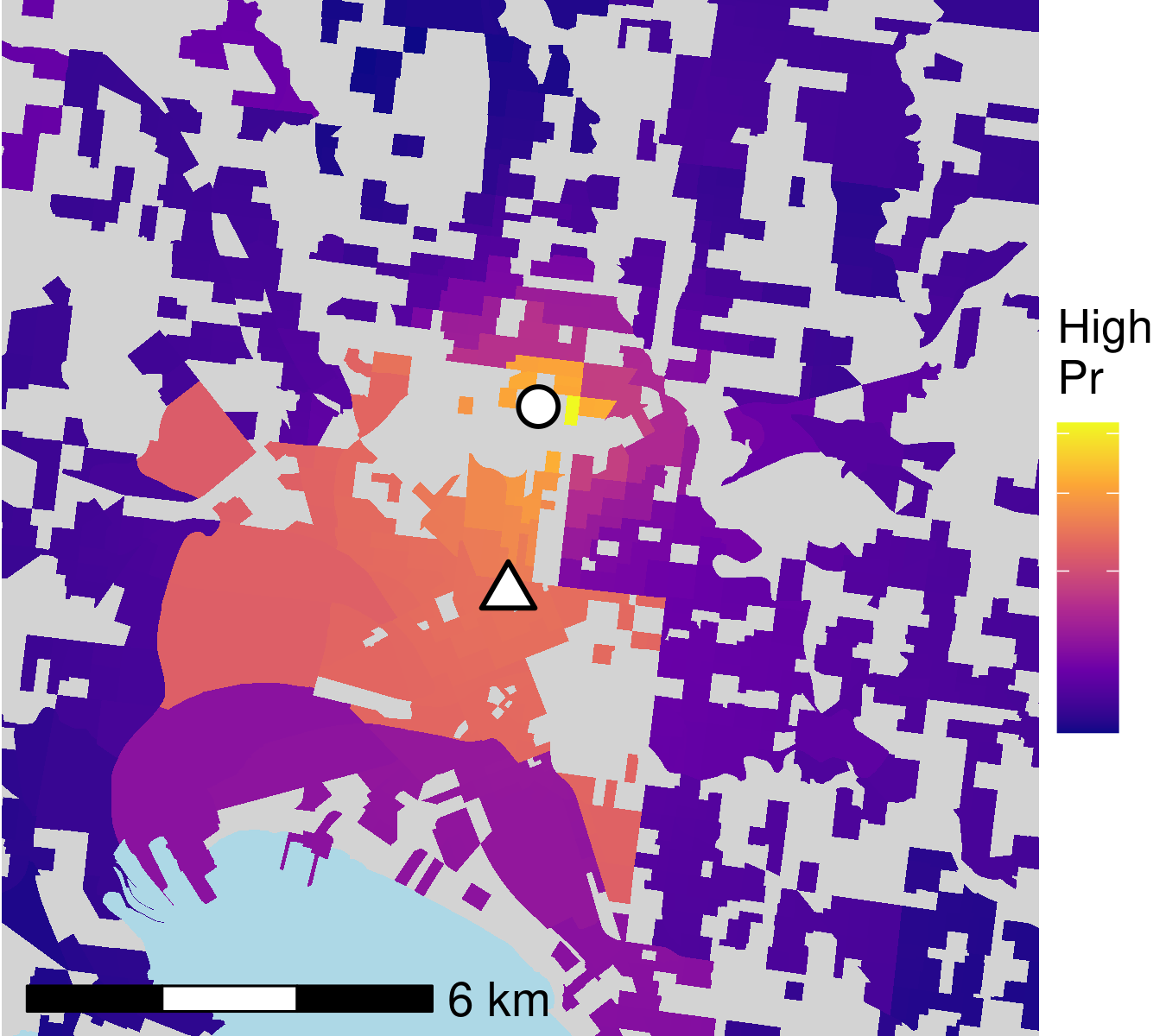}\\
			(e) & (f)\\
		\end{tabular}
		\caption{Selecting next region for a cycling trip from home (circle) to work (triangle) showing: region selection probability (Pr) for local and global distance distributions (a and b), region selection probability (Pr) for local and global destination attraction (c and d), number of trips (hop count) that would be reasonably required to reach home (e), and combined region likelihood (f).}
		\label{fig_selectingNextRegion}
	\end{figure}

	Selection of a destination region is handled by the getRegion function, which uses the current region, the location type of the destination region, the current mode, and the number of trips remaining until the home region is reached (hop count).
	Because trips with multiple stages tend to move further away from the home region, the remaining leg home can be unnaturally long.
	Hop count is used to filter candidate destination regions down to ones where it would be likely to for that mode, based on the number of trips remaining to get home assuming 95th percentile trip lengths based on their distance distribution.
	This ensures that the final trip home will not be unreasonably long, which can be a particular issue for walking trips. 
	The getRegion function is primarily a trade-off between selecting a destination that is a likely distance for the mode selected, and the attraction of the destination.
	Figure \ref{fig_selectingNextRegion} illustrates how the distance distribution and destination type probabilities are combined to create the probabilities needed to select the next region.
	
	%\begin{figure}[h]
	%	\centering
	%	%\includegraphics[width=1\textwidth]{virtualPopulation.pdf}
	%	\begin{tabular}{ccc}
		%		\includegraphics[width=0.31\columnwidth]{figs/1_distance.png} &
		%		\includegraphics[width=0.31\columnwidth]{figs/2_attraction.png} &
		%		\includegraphics[width=0.31\columnwidth]{figs/3_hop.png}\\
		%		(a) & (b) & (c)\\
		%		\includegraphics[width=0.31\columnwidth]{figs/4_global_distance.png} &
		%		\includegraphics[width=0.31\columnwidth]{figs/5_global_attraction.png} &
		%		\includegraphics[width=0.31\columnwidth]{figs/6_combined.png}\\
		%		(d) & (e) & (f)\\
		%	\end{tabular}
	%	\caption{Selecting next region for a cycling trip from home (circle) to work (diamond) showing: local and global distance probability (a and d), local and global destination attraction (b and e), number of trips that would be reasonably required to reach home, and combined region likelihood (f).}
	%	\label{fig_selectingNextRegion}
	%\end{figure}

	To provide a set of distances for the algorithm to choose from, an origin-destination matrix was calculated between the population-weighted centroids of \gls{abs} \gls{sa1} regions using the \gls{osm}-derived transport network.
	Population-weighted centroids were used as they are more representative of regions with uneven population distributions, and were calculated using the centroids of the \gls{abs} meshblocks with their 2016 census population.
	The \gls{sa1} centroids were then snapped to the nearest non-highway node and the shortest distance was calculated between all locations to populate the OD matrix.
	
	To calculate a distance distribution for each \gls{sa1} region, the population-weighted centroid was used as a base, selecting the closest 500 \gls{vista} trips for each transport mode. 
	A weighted log normal distribution was then calculated for each transport mode and centroid, recording their log-mean and standard deviation.
	The log-normal distribution was chosen as it better fit the distances than a normal distribution as there are no negative distances and can better exhibit a sharp peak at low distances regardless of transport mode.
	
	While it would be preferable to use a distance-based bandwidth as in the previous section to select trips to build the distributions, accurate representation of distributions requires more data than density-based measures.
	The closest 500 trips were chosen instead as only driving, with its 67,769 trips, was able to build representative distributions, whereas there were only 14,621 walking, 7,166 public transport, and 1,515 cycling trips.
	These distributions were calculated at the population-weighted centroids instead of the destination locations as the variation within the \gls{sa1} regions was not significant enough to warrant additional computation costs.
	
	Calculating the distance probabilities works by first filtering the OD-matrix to the current \gls{sa1} region, providing a set of distances to all other regions. 
	The log-mean and standard deviation for the chosen transport mode for the region are then used to filter potential destinations to only those with distances within the 5th and 95th percentiles of the log-normal distribution, ensuring unlikely distances are not selected.
	Potential destinations where the attraction probability was zero are also filtered to ensure that there would be a suitable destination location present.
	Probabilities are then calculated for the remaining destinations based on their distances and the log-mean and standard deviation for the nominated transport mode.
	These are then normalized so that their total equaled one.
	Because there are far fewer short distances than longer distances, the probabilities were further normalized by binning distances into 500 meter categories and then dividing each probability by the number of distances in that category.

	Destination attraction was calculated similar to mode probability, with the probabilities retrieved for the nominated destination category. Kernel Density Estimation (KDE) was again calculated at each candidate location for each destination category, using a weighted Gaussian kernel with a bandwidth of 3,200m for work, 300m for park, 200m for education, and 20,000m for commercial destinations. Then, these were aggregated to \gls{sa1} and normalized so that the total attraction value was equal to one for each category.
	
	Bandwidth selection was again based on calibration, where a variety of bandwidths were chosen, with the mode choice percentages aggregated to \gls{sa3} regions.
	These percentages were then compared with values obtained by aggregating the weighted \gls{vista} trips by destination category to \gls{sa3} region in order to select the bandwidth with the best fit for each destination category. It is noteworthy that park and education have relatively small bandwidths, indicating a high degree of local variation in attraction, whereas work, with its larger bandwidth, is less prone to local variation, instead more closely following the global trend of attraction being centered around the CBD. The commercial destinations required an even larger bandwidth, exhibiting even less deviation from the global trend.
	
	While these criteria ensure that regions are selected that are locally representative, it is also important to ensure that the transport demand is representative for all of Greater Melbourne.
	To account for this, distance distributions and destination attraction were tallied at the \gls{sa3} level, so that the destination region can also be selected based on how well it will improve the fit of the overall distributions.
	For example, consider Figure \ref{fig_selectingNextRegion}, where distances of previously generated cycling trips have been on average longer than the expected distribution.
	To account for this, the global distance probability is more likely to choose distances that are shorter than the local distance probability.
	This is also the case for destination attraction, where a higher probability is placed on the inner city as an insufficient number of work trips have arrived there.
	
	The final combined probability for selecting the next region is then calculated by filtering the candidate regions to those within the hop count, and then adding the local distance and destination attraction probabilities to the global distance and destination attraction probabilities, normalizing the probabilities so that their total equals one.
	From this point, a location could then be selected, as illustrated in Figure \ref{fig_selectingNextRegion}.

	In the final output of the algorithm (Table~\ref{tab:diary}), this step populates columns 
	\kk{SA1} (\gls{sa1} where this activity will take place), \kk{LocationType} (the type of location within the \gls{sa1} where the activity should take place, to be assigned in the subsequent step), \kk{Mode} (the travel mode by which the person will arrive at that activity; not applicable for the first activity of the day), and \kk{Distance} (the distance between the current and preceding \gls{sa1} regions derived from the OD-matrix).

	\subsection{Assigning locations to activities in statistical areas}\label{sec:6-place}
	
	With an \gls{sa1} region assigned for every stop, we now assign location coordinates based on the \gls{sa1} region and stop category.
	It is important to note that all home stops for an agent must share the same location.
	Locations are drawn from the set of candidate locations, which are points on the transport network that agents may move between.
	
	For the same destination category and \gls{sa1} region, certain locations will always be more popular than others.
	For example, an office tower will have more employees and, therefore, be a more attractive work destination than single building.
	To account for this, addresses were selected from Vicmap Address\footnote{\url{https://discover.data.vic.gov.au/dataset/address-vicmap-address}}, a geocoded database of property locations supplied by the Victorian government containing  2,932,530 addresses within the Greater Melbourne region.
	These were then assigned a land use category based on the meshblock within which they were located.
	In cases of meshblocks without any addresses within their boundaries, a single address was assigned at their geometries' centroid.
	
	In order to reduce the number of unique locations, the addresses were then snapped to the set of candidate locations mentioned in  Section \ref{sec:5-locate}, which are points on the transport network that agents may move between.
	The address counts were then used to create a selection probability by normalizing their number by \gls{sa1} region and destination category so that their total equaled one.
	This probability was then used to assign locations to each stop.
	
	In the final output of the algorithm (Table~\ref{tab:diary}), this step is responsible for populating columns 
	\kk{X} and \kk{Y}, representing the spatial coordinates of the location of activities, in the coordinate reference system of the input spatial data.
	
	\subsection{Assigning start and end times to activities}\label{sec:7-time}
	
	The final step of the algorithm converts the start and end times of activities, allocated initially at the coarse granularity of 30-min time bins (in Section~\ref{sec:3-plan}), into actual times of day. The main considerations in doing so are to 
	\begin{inparaenum}[(a)]
		\item add some random noise so that start/end times are sufficiently dispersed within the 30-minute duration of each time bin; and
		\item ensure that start/end times are ordered correctly so that activities do not end before their start, and activities do not start before the previous ones ends. This latter constraint becomes important where several starts and ends are being scheduled in the same time bin.
	\end{inparaenum}
	
	The method for achieving the desired time schedule is relatively straightforward. We first extract, for each person, the ordered vector of time bins corresponding to the sequence of start and end times for all activities. This is always a vector of even length, since every activity is represented by two sequential elements corresponding to its start and end. Further, this vector has values that increase monotonically, representing the progression of time as sequential activities start and end in the person's day. Next, this vector of time bin indices is converted to a time of day in seconds past midnight. We do this by taking the known start time of each time bin included in the vector and adding a randomly generated noise offset of a maximum duration of 30 minutes. This gives, across the population, start/end times evenly distributed within the time bins of activities. The obtained vector of times is then sorted in increasing order of numbers. This is necessary because the addition of random noise in the previous step can result in an out of order sequence of numbers within time bins where more than one activity is starting and/or ending. Sorting the vector rectifies any such issues and guarantees that the final ordering of start/end times are plausible and not mathematically impossible. Finally, the time values, which represent offsets in seconds past midnight, are converted to a more convenient \textsc{\textbf{hh:mm:ss}} 24-hour format.
	
	In the final output of the algorithm (Table~\ref{tab:diary}), this step is responsible for populating the columns 
	\kk{StartTime} and \kk{EndTime}, representing the start and end times of activities in the day.
	
	\section{Results}\label{sec_results}

	Section \ref{sec_cohort_activity_likelihoods} examines the activity propensities of the population cohorts. The following sections then compare a 10\% sample size population generated by our process to the real-world observations of the \gls{vista} travel survey to determine how closely our model matches the travel survey data distributions.
	To do so, the distributions of start, end, and duration of the activities (Section~\ref{sec_time_distributions}); distance distributions (Section \ref{sec_distance_distributions}); destination attraction (Section \ref{sec_destination_attraction}); and mode share (Section \ref{sec_mode_choice}) of the synthetic population were analyzed.
	Additionally, Section 
	\ref{sec_sample_size_accuracy} compares synthetic populations of varying sizes to determine the effect of sample size on accuracy.
	
	\subsection{Cohort activity likelihoods}\label{sec_cohort_activity_likelihoods}
	
	Table \ref{tab:cohort-activity-likelihood} shows the likelihood of activity choice for each of the five cohorts, along with their proportion of the total population.
	Cohort 1 is comprised of 0--19 year olds, with study as the main activity. Cohort 2 is the 20-24 year olds who mainly work, but have the next largest study proportion.
	Cohorts 3 and 4 are the working age adults, who mainly undertake working and shopping activities.
	Finally, cohort 5 represents the older population, who mainly undertake shopping and social/recreational trips.
	
	The purpose of creating these cohorts was to capture distinct patterns in travel behavior based on age and sex. Each cohort reflects unique activity tendencies, such as the higher study participation in younger cohorts or more frequent shopping trips among older adults. Without cohort segmentation, the entire population would be modeled using average activity patterns, missing these behavioral differences.
	
	The main outcome of the clustering process was the assignment of individuals to one of the five cohorts based on the probability of engaging in specific activities. As shown in Table \ref{tab:cohort-activity-likelihood}, each cohort had at least one activity that deviates significantly from the population average. This segmentation was to ensure more accurate activity chains and a realistic representation of travel behavior in the synthetic population.
	
	\begin{table}[]
		\centering
		\caption{Cohort activity choice likelihood}
		\label{tab:cohort-activity-likelihood}
		{\footnotesize\begin{tabular}{lrrrrrr}
				& \multicolumn{6}{c}{Cohort number (proportion of population)}                 \\ \cline{2-7} 
				& 1 (18.0) & 2 (6.3) & 3 (37.3) & 4 (24.5) & 5 (13.9) & Overall (100.0) \\ \hline
				Personal               & 5.5      & 8.3     & 6.6      & 9.2      & 17.8     & 8.7             \\
				Pickup/Dropoff/Deliver & 1.0      & 1.7     & 2.0      & 2.5      & 5.0      & 2.3             \\
				Shop                   & 17.9     & 18.8    & 26.6     & 39.2     & 38.6     & 29.3            \\
				Social/Recreational    & 27.3     & 24.0    & 19.6     & 18.8     & 27.9     & 22.2            \\
				Study                  & 45.6     & 10.9    & 1.4      & 1.0      & 0.4      & 9.7             \\
				Work                   & 2.7      & 36.3    & 43.9     & 29.2     & 10.3     & 27.7            \\ \hline
		\end{tabular}}
	\end{table}
	
	\subsection{Activity time distributions}\label{sec_time_distributions}
	
\begin{figure}[h]
		\centering
		\includegraphics[width=\textwidth]{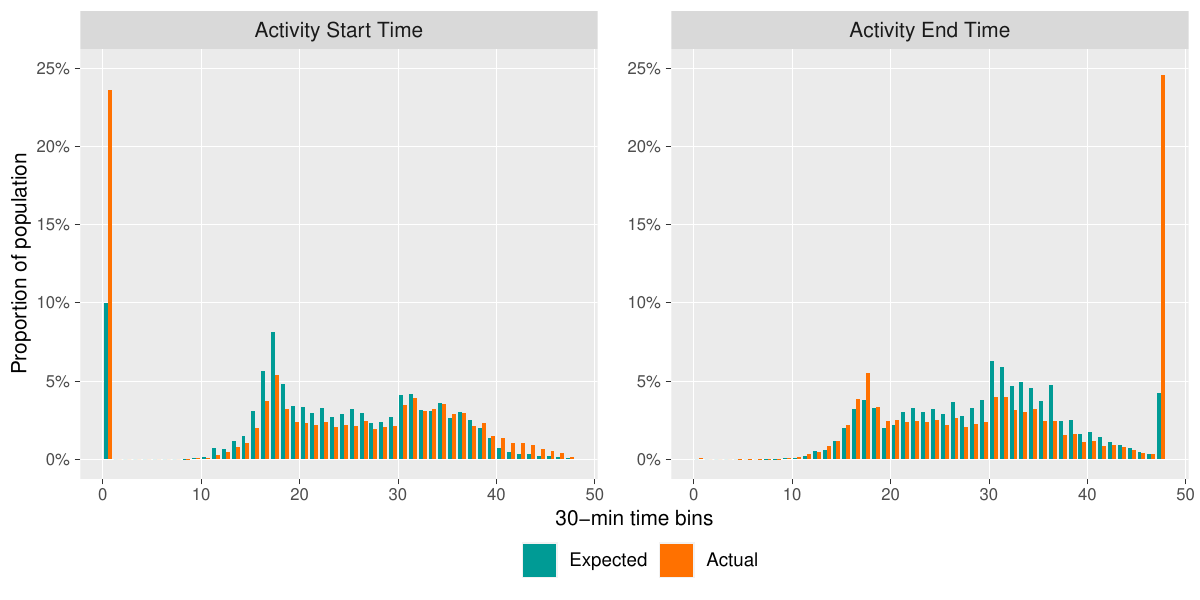}
		\caption{Start and end time histograms for activity chains generated by Algorithm~\ref{alg:plan-generation} (Actual) compared to the original \vista data (Expected), aggregated to 30-minute time bins. The exaggerated starts/ends are due to additional activities our algorithm inserts, to ensure that all synthetic agents always start and end their day at home.}
		\label{fig:time-histograms}
	\end{figure}

	Figure \ref{fig:time-histograms} shows the distribution of start (end) times of all activities of the synthetic population generated by Algorithm~\ref{alg:plan-generation} (Actual), compared with the original \vista data (Expected).
	%, aggregated to 30-minute time bins of the day.
	%
	The x-axis shows the time of day segmented into 48 time bins of 30-minutes duration, and the y-axis shows the proportion of the entire population that started (ended) activities in each time bin.
	
	The start (end) times of the activities in general matched the desired distributions of \vista. A morning peak is evident around 8–9am (bins 16–18) when a significant proportion of the population starts work, study, school drop-offs, and shopping activities. Towards the end of the day, a second longer evening peak can be seen from 3–6pm (bins 30–36) from school pick-ups and trips to shops, and as a large proportion of the population finishes work and transitions to more social and recreational activities.
	
	A notable distortion in the generated distributions was visible at the start and end of the day in Figure \ref{fig:time-histograms}. This was caused by artificial 'home' activities that our algorithm inserts for any generated activity chains that do not begin and/or end at home. This is not strictly necessary, but was done to ensure compatibility with certain target downstream simulation tools such as MATSim~\citep{horni_multi-agent_2016}. Evidently, these exaggerated peaks at the start (end) of the day caused an overall drop in the amplitude of the peaks during the remainder of the day. Ignoring artificially distorted bins at the extreme ends of the day, the relative shape of the generated distribution was generally a good match to the \vista data.
	%Large times at start and end of day are artificial, and are a result of our algorithm requiring all activity chains to start and end with home activities. Otherwise fits extremely well. Can see the morning and evening peaks.
	
	\begin{figure}[h]
		\centering
		\includegraphics[width=\textwidth]{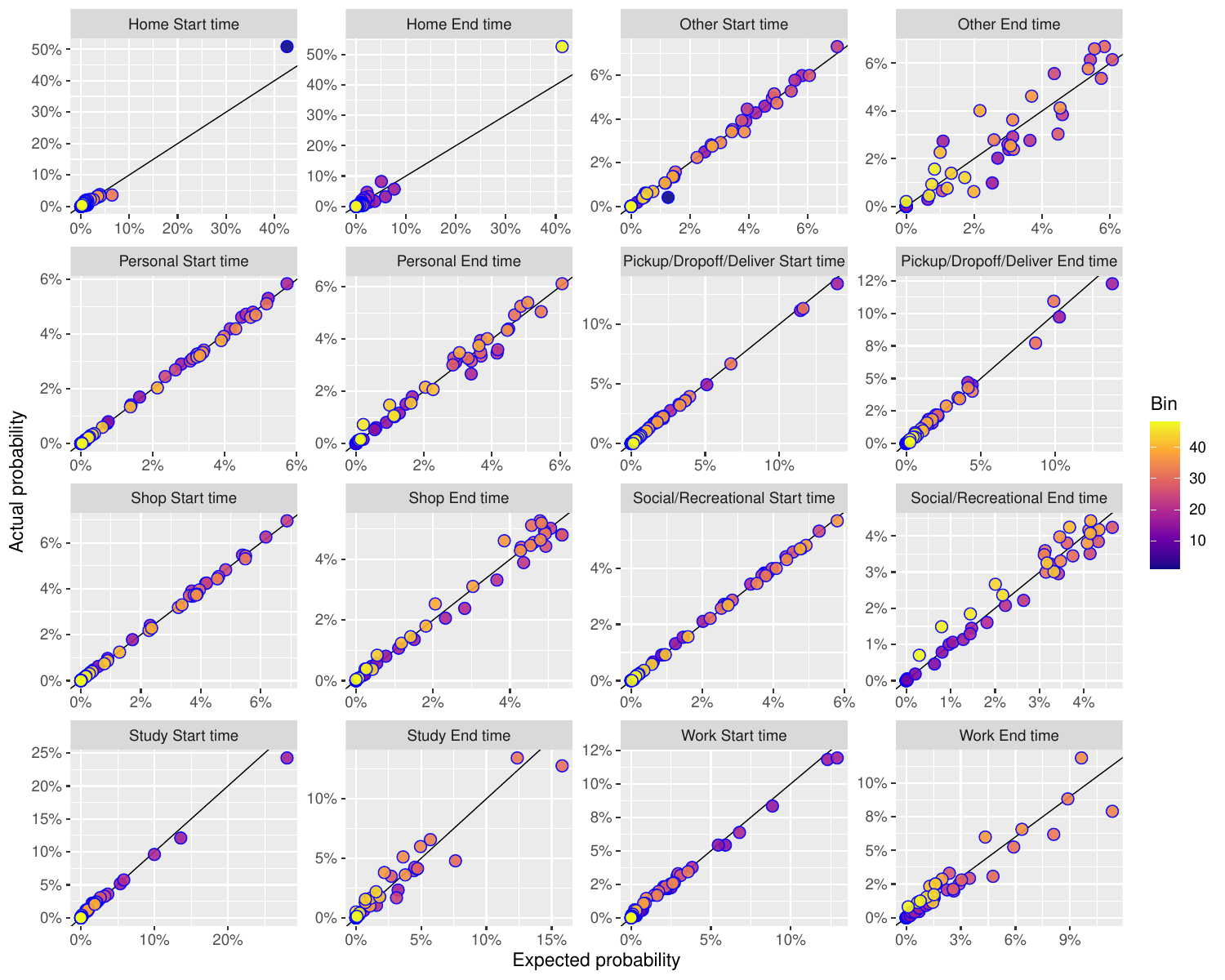}
		\caption{Expected versus actual start and end time distributions by activity.}
		\label{fig:time-activities}
	\end{figure}
	
	A more detailed view of the different types of generated activities is given in Figure \ref{fig:time-activities}, which shows the breakdown of activity start and end time probabilities. As before, results are aggregated to 30-minute time bins and compared to \vista.
	On the x-axis is the proportion of the population starting (ending) a certain activity type in \vista, and on the y-axis is the proportion of the synthetic population starting (ending) the same type of activity. The dots represent the time of day (48 bins). The solid line is a guide that shows where the dots should fall if the generated distribution was a perfect match to the desired (\vista) distribution.
	
	Figure~\ref{fig:time-activities} shows that the output activities fit the desired \vista distributions extremely well, considering that the task was to generate a series of activity chains such that, aggregated, the activities (x8) and their start (end) times (x2), across the day (x48), all (i.e., 8*2*48=768 considerations) matched the target distributions. Start times across all activities and time bins generally fit very well compared to end times. This is because end times were generated based on start times, which resulted in some compounding of errors.
	
\begin{figure}[h]
		\centering
		\includegraphics[width=\textwidth]{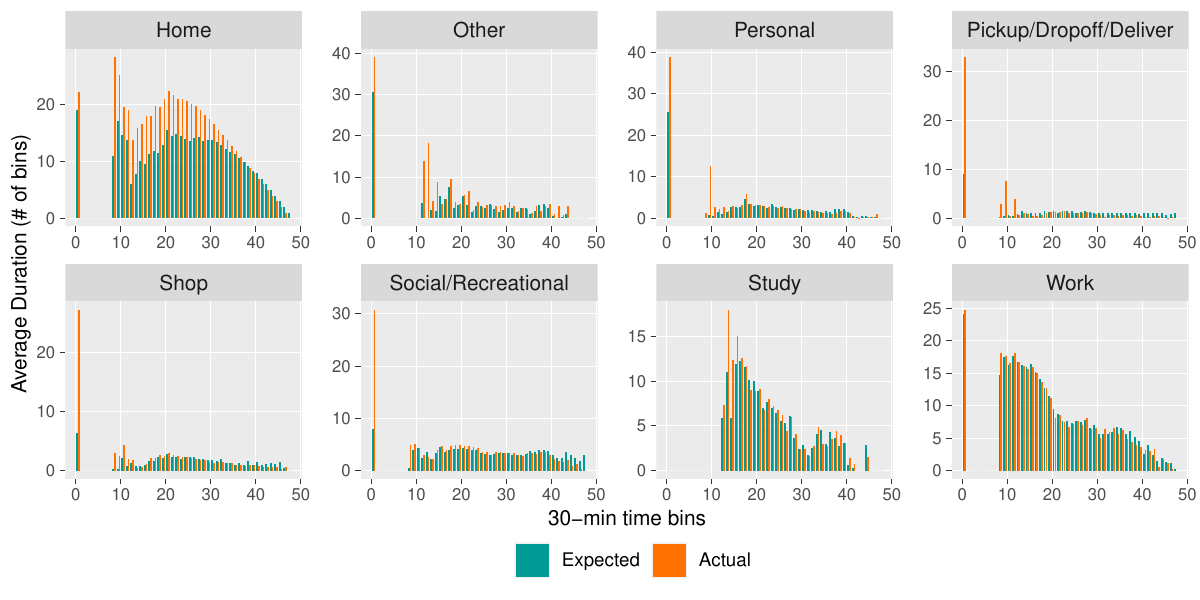}
		\caption{Activity duration histograms, aggregated to 30-minute time bins.}
		\label{fig:duration-histograms}
	\end{figure}

	Finally, the activity chains generated using Algorithm~\ref{alg:plan-generation} should also match the durations of the activities observed in \vista (i.e., an additional 8*48=384 considerations). Figure \ref{fig:duration-histograms} shows how the algorithm (Actual) performs against \vista (Expected) for each activity type. 
	
	The x-axis shows activity split across the 48 time bins, and the y-axis shows the duration of the activity when started in each time bin. The generated activities' durations matched the desired durations well, from long activities such as work, to short ones related to pick-ups and drop-offs. Some distortion is evident in the first time bin of the day across most activities, and for longer duration Home activities during the day.

	Overall, the generated activities were a good fit to the desired \vista distributions in all (384 + 768 = 1152) considerations, excluding outliers at the extreme ends of the day. It is worth noting that the self-correcting nature of Algorithm~\ref{alg:plan-generation}, which continually steers activity generation towards under-represented and away from over-represented activity-bin combinations, ensures that the accuracy of the generated distributions continues to improve with the number of generated activity chains. The results presented here were based on a 10\% sample of the entire population of Greater Melbourne. Section \ref{sec_sample_size_accuracy} gives details on how the accuracy improves with sample size, along with guidance on the recommended minimum sample size for downstream use of the generated population.
	
	\subsection{Distance distributions}\label{sec_distance_distributions}
	Given that the synthetic population created by this work provides locations for each agent's destination, but not routing information, distances were instead calculated based on the distance between the \gls{sa1} regions provided by the OD-matrix of Section \ref{sec:5-locate} (i.e., the shortest path distance along the road network between the population-weighted centroids of the two \gls{sa1} regions).
	To ensure consistency, the distances from the \gls{vista} trips dataset were also replaced with the distance between \gls{sa1} regions.
	
\begin{figure}[h]
		\centering
		\includegraphics[width=\textwidth]{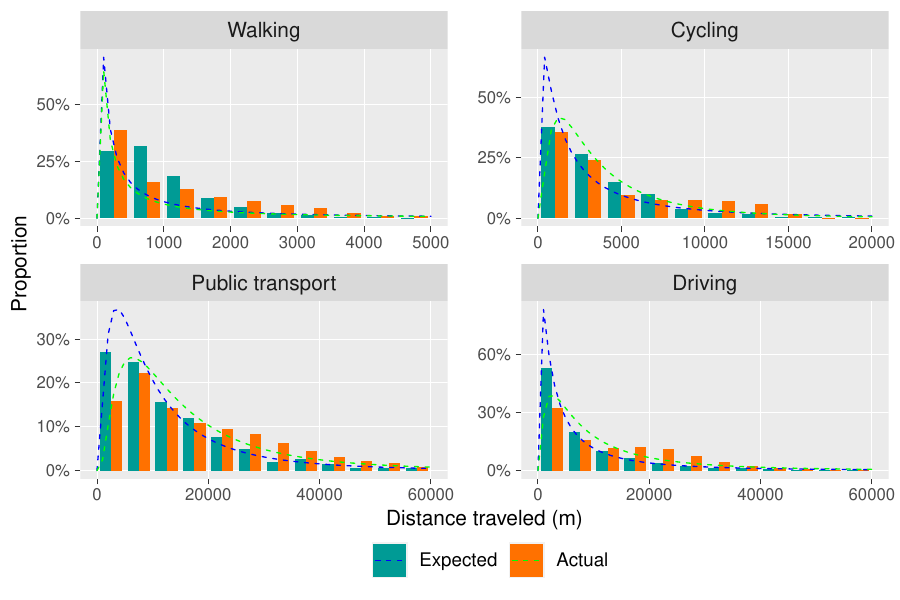}
		\caption{Distance histograms. Dashed lines represent the fitted log-normal distributions.}
		\label{fig:distance-histograms}
	\end{figure}

	Figure \ref{fig:distance-histograms} shows the distance distributions generated by the synthetic population (Actual) plotted alongside the weighted original \vista data (Expected) for the four transport modes.
	Fitted log-normal distributions are also plotted as dashed lines.
	In general, the actual distributions matched the expected distributions closely, although the actual distributions appear to have larger values in the longer distances.
	
\begin{figure}[h]
		\centering
		\includegraphics[width=\textwidth]{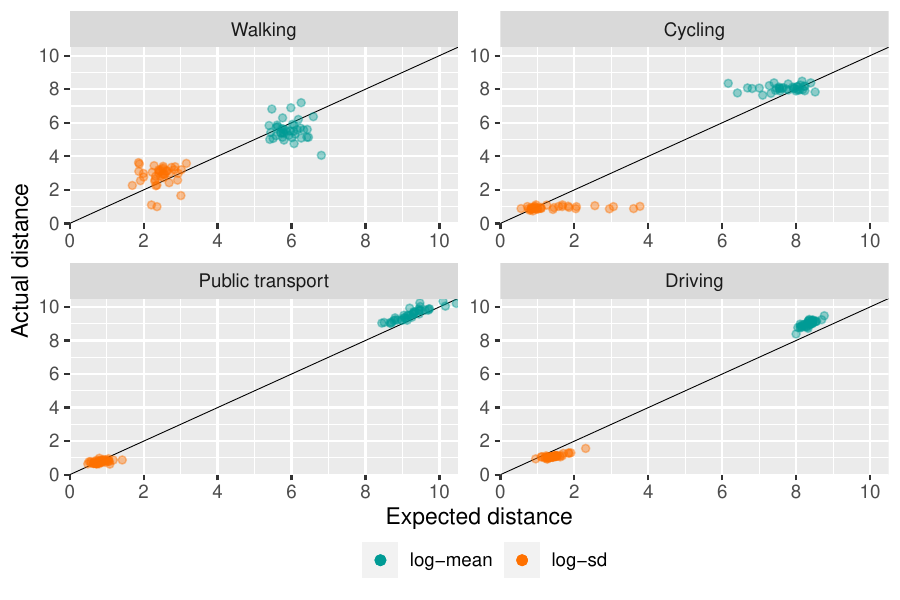}
		\caption{Expected versus actual distance distributions aggregated to \gls{sa3} regions in km.}
		\label{fig:distance-distributions}
	\end{figure}

	To determine whether the spatial variation of the distance distributions was captured, actual and weighted expected distance distributions have been aggregated to the 40 \gls{sa3} regions comprising Greater Melbourne.
	Figure \ref{fig:distance-distributions} shows the log-normal and standard deviation of these distance distributions, with each dot representing an \gls{sa3} region.
	
	Again, the solid line is a guide showing where the dots should fall if the generated distribution were a perfect match to the desired (VISTA) distribution.
	Of the four modes, cycling appeared to have the largest variation, which is expected given that it is based on only 1,515 cycling trips from the \vista travel survey.
	While walking had similar expected and actual values, there was little positive correlation.
	This is reasonable given that how far people are willing to walk, likely has little to no spatial variation.
	In contrast, public transport and driving showed a much stronger positive correlation, meaning that the spatial variation of these modes is being captured.
	Additionally, the actual log-mean is on average smaller than the expected log-mean for driving and public transport, whereas the log-standard deviation was larger.
	This is consistent with Figure \ref{fig:distance-histograms} given these modes tend to have a lower peak and longer tail in their histograms.
	
	\subsection{Destination attraction}\label{sec_destination_attraction}
\begin{figure}[h]
		\centering
		\includegraphics[width=\textwidth]{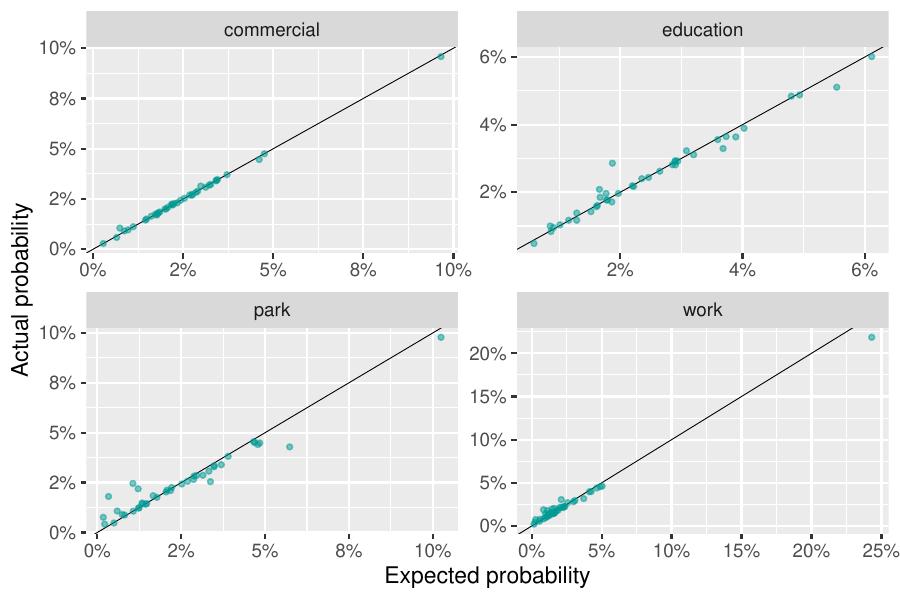}
		\caption{Expected versus actual destination probabilities aggregated to \gls{sa3} regions.}
		\label{fig:destination-attraction}
	\end{figure}

	In order to determine if the spatial variation of the destination activities was captured, Figure \ref{fig:destination-attraction} compares actual and weighted expected destination probabilities aggregated to \gls{sa3} regions.
	It is important to note that the spatial variation of home locations has not been plotted as the number of home locations for each \gls{sa1} is a value that is explicitly used when generating the synthetic population.
	
	It is also important to note that for commercial, park, and work activities, one of the \gls{sa3} regions had a much higher chance of being selected as it is the region containing Melbourne's \gls{cbd}.
	In general, \gls{sa3} regions had similar expected and actual probabilities regardless of destination type, indicating a good fit for destination selection.
	The park destination type did however display larger variation, likely due to being a comparatively unpopular activity and therefore having fewer trips.
	This was also the case to a lesser extent for education activities.
	Additionally, all destination types show a large variance in probabilities along with a positive correlation, indicating that the spatial variation has been represented.
	
	\subsection{Mode share}\label{sec_mode_choice}
\begin{figure}[h]
		\centering
		\includegraphics[width=\textwidth]{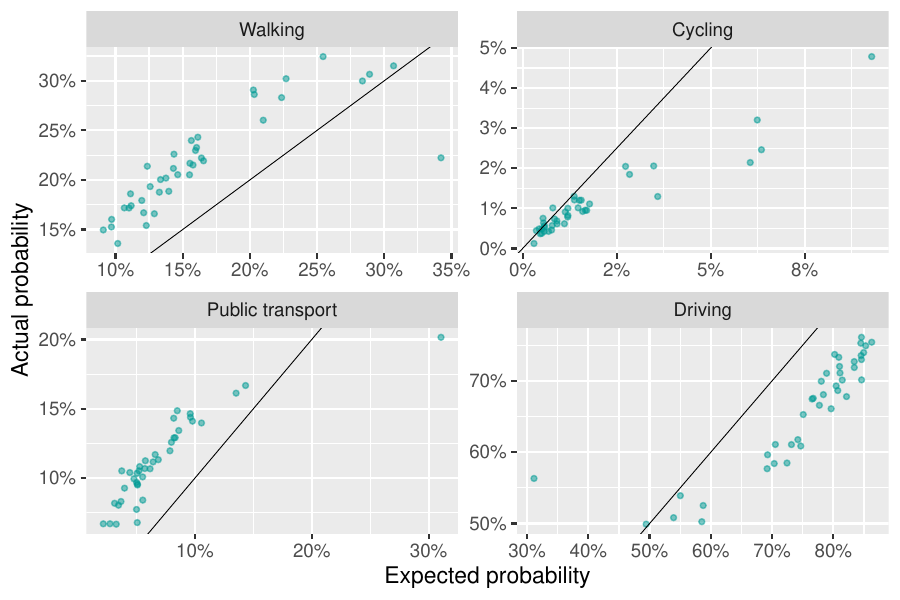}
		\caption{Expected versus actual mode share proportions aggregated to \gls{sa3} regions.}
		\label{fig:mode-choice}
	\end{figure}

	In order to determine if the spatial variation of the transport mode share is being captured, Figure \ref{fig:mode-choice} compares actual and weighted expected mode share proportions that are aggregated to \gls{sa3} regions.
	
	Walking, public transport, and driving were represented accurately, with very similar expected and actual probabilities baring one outlier.
	Additionally, the variance and positive correlation indicate that the spatial variation has been represented.
	The outlier present for these three modes was again the \gls{sa3} region containing the \gls{cbd}.
	Specifically, driving has been overrepresented in the \gls{cbd}, causing the other regions to be underrepresented.
	Likewise, this has caused walking and public transport to be underrepresented in the \gls{cbd}, and therefore overrepresented in the other regions.
	Cycling was a comparatively unpopular transport mode but expected and actual values are similar.
	There was also a moderate amount of variance in probabilities and a positive correlation, indicating that the spatial variation has at least been represented, although not as accurately as the other modes.

	\subsection{Sample size accuracy}\label{sec_sample_size_accuracy}
	
	So far, the results have been calculated using a 10\% sample population, as that is a common size used in \gls{abm} simulations. However, it is important to determine at what sample sizes a synthetic population will be representative of the underlying \gls{vista} travel survey.
	
	\begin{table}[h]
		\centering
		\caption{Average difference between actual and expected results for various sample sizes.}
		\label{tab:sample-size-table}
		{\footnotesize
			\begin{tabular}{llrrrr}
				&                  & \multicolumn{4}{c}{Population sample size}                   \\ \cline{3-6} 
				&                  & \textbf{0.1\%} & \textbf{1\%} & \textbf{5\%} & \textbf{10\%} \\ \hline
				\multirow{4}{*}{Destination attraction} & Commercial       & 0.52\%         & 0.17\%       & 0.06\%       & 0.04\%        \\
				& Education        & 0.63\%         & 0.25\%       & 0.14\%       & 0.12\%        \\
				& Park             & 0.72\%         & 0.34\%       & 0.29\%       & 0.29\%        \\
				& Work             & 0.67\%         & 0.32\%       & 0.29\%       & 0.28\%        \\ \hline
				\multirow{4}{*}{Mode share}     & Walking          & 7.27\%         & 8.68\%       & 7.06\%       & 6.07\%        \\
				& Cycling          & 1.15\%         & 0.88\%       & 0.79\%       & 0.73\%        \\
				& Public transport & 5.59\%         & 5.96\%       & 5.26\%       & 4.73\%        \\
				& Driving          & 12.24\%        & 13.94\%      & 11.64\%      & 10.24\%       \\ \hline
			\end{tabular}
		}
	\end{table}
	
	Table \ref{tab:sample-size-table} shows the average difference between the actual results generated by the synthetic populations and weighted expected results from the \vista trips dataset aggregated to \gls{sa3} regions for both destination attraction and mode share. In general, there was a clear trend towards accuracy with increasing sample size. This is to be expected as the \textit{sample} stage (Section \ref{sec:3-plan}) generates activity chains that better fit the \gls{vista} travel survey.
	Likewise, the global distance distribution and destination attraction of the \textit{locate} stage (Section \ref{sec:5-locate}) also fit their choices to the distributions of the travel survey.
	For both of these sections, each new trip represents a chance to better fit their distributions, larger sample sizes should produce increasingly representative results.
	
	It is noteworthy that walking, public transport, and driving were less accurate for the 1\% sample than the 0.1\% sample, suggesting that the results are not stable at these sample sizes, and that larger sample sizes should be used if a representative sample is required. 
	Additionally, the gains in accuracy diminish with increasing sample size, with little accuracy gain between the 5\% and 10\% samples.
	Therefore, we recommend that a minimum sample size of 5\% be used, but ideally 10\% should be used in agent-based modeling simulations.
	
	\section{Discussion}\label{sec_discussion}
	
	In this paper, we presented an algorithm to create activity-based transport demand, suitable for use as input to agent-based models, using a combination of clustering, probabilistic, and gravity-based approaches. 
	While this work specifically focused on the Greater Melbourne region, our method is completely open and replicable, requiring only publicly available data.
	It is important to note that we did make use of non-public \gls{vista} data in order to achieve a finer spatial accuracy of the mode and destination choices than would have been achievable with the public data.
	We have, however, made the derived products necessary for the activity-based transport demand model available.
	While our work specifically made use of the \gls{vista} travel survey and population demographics from the \gls{abs} Census, similar datasets are available in many regions in Australia and also globally, meaning that adapting this process to other jurisdictions would be a straightforward task.
	
	The first innovation produced by our hybrid model was to dispense with the cloning of preexisting activity chains from the travel survey and instead generate individual activity chains for every agent, tailored to their cohort.
	It is important to note that the activity chains we generate do not rely on the high-level structure of a tour-based approach, where a pre-defined range of activities are assigned to specific tour components (e.g., work activities to work tours).
	Our activity scheduling model dispenses with this approach, instead allowing activities to be selected based on their type, start time, and duration without the inherent rigidity of tour-based models. This lack of a higher level structure constraining the activities instead allows a greater variance of tours to instead form organically.
	
	While cloning activity chains is compatible with splitting a travel survey population into cohorts based on behavior, it requires that there are sufficient trips within each cohort.
	Specifically, there is a limit on what sort of cohorts can be generated, as when there are fewer travel survey participants in a cohort, any anomalous behavior is at higher risk of being duplicated.
	This is a particular issue for active transport modes such as walking and cycling, as these trips are typically underrepresented when compared to driving and public transport.
	By moving from cloning activity chains, to converting them into activity distributions that any number of representative activity chains may be generated from, we ensured that our method does not rely on the accuracy and replication of individual trips.
	Additionally, our algorithm is self-correcting, improving its accuracy as more synthetic agents are added to the population.

	The second innovation presented in this work was to add a spatial context to the selection of destinations used by agents. Specifically, local probabilities for mode choice, trip distance, and destination attraction were generated and calibrated for each of the 10,289 \gls{sa1} regions within Greater Melbourne.
	While this approach ensures that local variation is represented in the transport demand model, ensuring that the distance distributions of trip lengths and the activity-based attraction of destination locations are both accurately represented is a balancing act.
	By altering the weights of the models contributing to destination selection, it is trivial to create an \gls{abm} that almost perfectly fits either the distance distributions or the destination activities in a local or global context.
	Ultimately, the weights chosen were a compromise to ensure that each of these factors were sufficiently accurate to allow agents to better consider trip length and destination location.
	Specifically, destination attraction was given a higher weighting than the distance distributions to ensure that sufficient agents were traveling into the \gls{cbd}, which has caused a greater number of trips of larger distances. 
	
	Our final innovation was to incorporate a hop-count measure to filter candidate destination regions. 
	By taking into account the number of trips remaining for an agent, we can ensure that they do not select a destination that would be unreasonable to return home from, given their transport mode.
	This is a particular issue for activity chains with several trips as they tend to move further away from the home region, which can potentially cause the final trip home to be unnaturally long.
	Active transport modes, such as walking and cycling, are more susceptible to this, as their distance distributions are much shorter than public transport or driving.
	Specifically, distance is a key factor for most mode choice algorithms, meaning that a walking activity chain with a long trip home will score poorly on its final trip potentially forcing a shift in modes for the entire chain.
	If enough of these anomalies are present, this would disproportionately reduce the number of agents utilizing active transport modes such as walking and cycling.
	
	To summarize, this study developed an algorithm based directly on travel activity patterns from a travel survey enabling more realistic generation of travel demand without replication and without constraint rather than directly modelling components for mode-choice through econometric techniques such as nested-logistic regression models that are reliant on tour components (i.e., individual trips within the tour).  Our approach produces realistic tour components as an emergent property within the activity pattern generation. Our approach is more flexible; less computationally burdensome; and, follows calls from other authors to explore techniques with reduced data requirements \citep{nayak_critical_2022}. 
	
	The main limitation of our approach is that it can only build a realistic baseline set of activity patterns and it is insensitive to changes in certain factors such as number of activities and destination choice although future research could explore ways of incorporating these. We can change the travel time and the travel mode, as demonstrated in our previous research \citep{jafari_activity-based_2024}, whilst our synthetic population is sensitive to mode and timing changes.
	
	One key limitation of the current model is that driver and passenger trips are treated independently under a single ``car'' mode, meaning that shared travel, such as carpooling or family members traveling together, has not been explicitly captured. As a result, the model may underrepresent some coordinated travel behaviors and household-level interactions. Addressing this limitation is a future avenue of research where trip chains are generated at the household level, enabling better representation of driver-passenger interactions and shared mobility patterns.
	
	Additionally, car ownership is only indirectly incorporated through spatial location, where areas located further from the city-center with poor public transport accessibility have higher car ownership than more established city-center areas, where public transport accessibility is usually better \citep{gunn_early_2020}. A more complex car ownership model was not developed for this reason, and because there is high car ownership and preference for driving evident in the characteristics and travel behaviors of participants in the \gls{vista} travel survey.
	
	Future work could involve further enhancements to cohort generation and population matching so that demographics other than age and gender, such as income and household-level auto-ownership, could be considered when generating activity chains.
	Presently, our \gls{abm} is compatible with agent-based simulations such as \gls{matsim}, but does not yet make use of its capabilities in refining the model.
	Specifically, agents in our \gls{abm} base their schedule timings and choice of transport modes on information derived from the \gls{vista} travel survey.
	By simulating these trips within an agent-based model, interactions between agents can produce realistic traffic congestion, which can enable improved trip timings. However, further validation of the congestion outputs is necessary to confirm their realism and will be part of future work.
	This provides the opportunity to create an iterative process in future work, where the generated transport demand from our model could be used as input for agent-based simulations such as MATSim. The outputs of such simulations—such as congestion levels and travel times—can then feed back into the demand model to refine activity chains and schedules. This iterative process would improve the accuracy of both demand generation and transport modeling. 
	
	Additionally, if this iterative workflow, where a simulation model tests interventions and feeds results back to the demand generation model, is implemented, it could assess the impact of various policy and infrastructure interventions not only on mode choices and traffic flows but also on travel itineraries and timing. For example, the model could simulate the effects of reducing speed limits, which in the simulation might result in shifts toward public transport and cycling. This shift, when fed back into the demand generation model, could lead to the selection of new destinations that align better with the updated travel modes, thereby changing the sequence and locations of subsequent activities. Such an iterative process enables a more detailed understanding of the implications of interventions, capturing both direct and secondary effects on travel behavior and network performance.
	
	In conclusion, the process presented in this work was able to successfully generate an \gls{abm} with the demographic attribute of the \gls{abs} Census combined with travel behavior characteristics from the \gls{vista} travel survey that included information from activity type, start and end times, and duration, along with mode and destination choice, and distance distribution.
	
	\section*{Acknowledgments}
	This paper is part of the development of a city-scale MATSim model for Melbourne. We thank the MATSim Melbourne community and our colleagues at Healthy Liveable Cities Lab, Centre for Urban Research, RMIT University, for their inputs and comments that informed this research. 
	
	\section*{Funding}
	LG and AB’s time on this project is funded by the NHMRC funded Australian Prevention Partnership Centre (\#9100001); and BGC is supported by an RMIT VC Professorial Fellowship. AJ is supported by an Australian Government Research Training Program Scholarship. DS’s time on this project is funded by Collaborative Research Project grants from CSIRO’s Data61, Australia (2018-19, 2020-21, 2022-23). 
	Data Availability Statement:
	The data presented in this study and the code used to generate the data are openly available at \url{https://github.com/matsim-melbourne/demand}.
	
	\section*{Author Contributions} 
	Alan Both: Conceptualization, Formal analysis, Investigation, Methodology, Software, Validation, Visualization, Writing - original draft; Writing - review \& editing. Dhirendra Singh: Conceptualization, Formal analysis, Investigation, Methodology, Software, Validation, Visualization, Writing - original draft; Writing - review \& editing. Afshin Jafari: Conceptualization, Formal analysis, Investigation, Methodology, Software, Validation, Writing - original draft; Writing - review \& editing. Billie Giles-Corti: Conceptualization, Funding acquisition, Writing - original draft; Writing - review \& editing. Lucy Gunn: Conceptualization, Writing - original draft; Writing - review \& editing.

	\bibliographystyle{apalike}
	\bibliography{references.bib}
	
\end{document}

%% file: figs/vista-trip.tex
\begin{tikzpicture}[->,>=stealth',shorten >=1pt,auto,node distance=3cm,semithick]
  \tikzstyle{every state}=[fill=black!05!white,draw=black!30!white,text=black,minimum width=1.5cm]

  \node[state]         (A)              {Home};
  \node[state]         (B) [left of=A]  {Work};
  \node[state]         (C) [left of=B]  {Buy};
  \node[state]         (D) [right of=A] {Buy};

  \path (A) edge [bend left]  node {1} (B)
        (B) edge [bend left]  node {2} (C)
        (C) edge [bend left]  node {3} (A)
        (A) edge [bend left]  node {4} (D)
        (D) edge [bend left]  node {5} (A);
\end{tikzpicture}

%% file: figs/matrix.tex
\begin{tikzpicture}
  \tikzset{
    cell/.style={
      black!70,
      draw=black!20,
      anchor=north west,
      rounded corners,
      minimum width=0.65cm,
      minimum height=0.65cm}
  }
  \tikzset{
    acell/.style={cell,fill=black!05}
  }
  \tikzset{
    bcell/.style={cell,fill=white,text=red,minimum width=0.95cm}
  }

    \node[draw=black!20,
      fill=black!10,
      anchor=north west,
      rounded corners,
      minimum width=4.925cm,
      minimum height=3.1cm]
      at(2.0,-3) {};

 \draw[
    red,
    decorate,
    decoration={brace,amplitude=5pt},
    xshift=0pt,yshift=0pt]
    (2,-2.8) -- (6.9,-2.8)
    node [red,midway,yshift=12pt]{\T};
  \draw[
    red,
    decorate,
    decoration={brace,mirror,amplitude=5pt},
    xshift=0pt,yshift=0pt]
    (1.8,-3) -- (1.8,-6.1)
    node [red,midway,xshift=-12pt]{$\vert\mathcal{A}\vert$};

  \draw[black,dotted] (3.0,-3) -- (3.0,-6.1);
  \draw[black,dotted] (3.975,-3) -- (3.975,-6.1);
  \draw[black,dotted] (4.955,-3) -- (4.955,-6.1);
  \draw[black,dotted] (5.935,-3) -- (5.935,-6);

  \node[anchor=north west] at(5.1,-3.7){$\ldots$};
  \node[anchor=north west] at(5.3,-4.3){$\vdots$};
  \node[anchor=north west] at(5.1,-5.6){$\ldots$};

  \node[bcell] at(2.025,-3.025) {$p_{11}$};
  \node[bcell] at(3.000,-3.025) {};
  \node[bcell] at(3.975,-3.025) {};
  \node[bcell] at(5.945,-3.025) {};

  \node[bcell] at(2.025,-3.725) {};
  \node[bcell] at(3.000,-3.725) {};
  \node[bcell] at(3.975,-3.725) {$p_{23}$};
  \node[bcell] at(5.945,-3.725) {};

  \node[bcell] at(2.025,-5.425) {};
  \node[bcell] at(3.000,-5.425) {};
  \node[bcell] at(3.975,-5.425) {};
  \node[bcell] at(5.945,-5.425) {\scriptsize$p_{\vert\mathcal{A}\vert\mathcal{T}}$};

\end{tikzpicture}

%% file: figs/algo-step3.tex
\begin{algorithm}[h]
\scriptsize
\SetKwData{a}{a}
\SetKwData{i}{i}
\SetKwData{j}{j}
\SetKwData{s}{s}
\SetKwData{p}{p}
\SetKwData{r}{r}
\SetKwData{TRUE}{TRUE}
\SetKwData{arch}{a.archetype}
\SetKwFunction{Colnames}{Colnames}
\SetKwFunction{GetSignature}{GetSignature}
\SetKwFunction{GetSignatureRow}{GetSignatureRow}
\SetKwFunction{GetColumn}{GetColumn}
\SetKwFunction{Sum}{Sum}
\SetKwFunction{CumSum}{CumSum}
\SetKwFunction{Runif}{Runif}
\SetKwFunction{Match}{Match}
\SetKwInOut{Input}{Input}\SetKwInOut{Output}{Output}
\Input{Activity start time distribution matrix \Ds, activity end-time distribution matrix \De, number of time bins \T, number of trip chains to generate \N}
\Output{Set of generated trip chains \C}
\BlankLine
\tcp{Init list \C for storing generated trip chains}
\tcp{Init matrix \Dso for storing start-time distributions of chains in \C}
\tcp{Init matrix $\Delta$ for the difference from expected start time distributions}
\For{$n\leftarrow 1$ \KwTo \N}{
    \tcp{normalise to 1 all rows in $\Delta$ that have non-zero values}
    $\Delta\leftarrow\mathcal{D}_s-\Delta$\tcp*{get the difference from expected}
    \tcp{make $\Delta$ rows $\geq 0$ by adding the minimum value of each row as offset}
    \tcp{for every zero-cell in \Ds make the corresponding cell in $\Delta$ zero }
    \tcp{re-normalise to 1 all rows in \Dso that have non-zero values}
    $b \leftarrow 1$ \tcp*{start at time bin 1}
    \tcp{Init list $\Phi$ for storing a new trip chain}
    \While(\tcp*[f]{while not at the final time bin}){$b < $\T}{
       $p\leftarrow$\Sum{$\mathcal{D}_s[,b]$}/\Sum{\Ds}\tcp*{get \% of activities that should start in $b$}
        $\acute{p}\leftarrow$\Sum{$\acute{\mathcal{D}_s}[,b]$}/\Sum{\Dso}\tcp*{get \% of activities starting in $b$ so far}
        \If(\tcp*[f]{have more than the required share of starts for $b$}){$\acute{p}>=p$}{\tcp{increment b and skip to start of loop}}
        $\delta\leftarrow\Delta[,b]$\tcp*{pick the difference probabilities for time bin b}
        \tcp{ set all zero-cells in $\delta$ that are non-zero in $\mathcal{D}_s[,b]$ to 0.001}
        \tcp{if all $\delta$ cells are zero then increment b and go to start of loop}
        \tcp{normalise $\delta$ to 1}
        \tcp{use $\delta$ values to probabilistically select an activity $a$ from \A}
        $b_s \leftarrow b$\tcp*{set the start bin for activity $a$}
        \tcp{from \De get the row $\gamma$ for activity $a$ starting in bin $b_s$}
        \tcp{if $\gamma$ has all cells zero, then set all cells to 1}
        \tcp{use $\gamma$ values to probabilistically select an end bin $b_e$ for $a$}
        \tcp{append list $[a,b_s,b_e]$ to the end of $\Phi$}
        $b \leftarrow b_e$\tcp*{skip to the bin that activity $a$ finished on}
    }
    \tcp{if $\Phi$ is empty then append list $[Home,1,\mathcal{T}]$ to it }
    \tcp{collapse consecutive blocks of the same activity in $\Phi$ into one}
    \tcp{append $\Phi$ to the end of \C}
    $\acute{\mathcal{D}_s}[a,b_s]\leftarrow\acute{\mathcal{D}_s}[a,b_s]+1$\tcp*{increment the count for $a$ starting in $b_s$}
}
\caption{Algorithm for generating travel survey like activity chains}
\label{alg:plan-generation}
\end{algorithm}

%% file: figs/algo-step5.tex
\begin{algorithm}[H]
\scriptsize
\SetKwData{TRUE}{TRUE}
\SetKwData{FALSE}{FALSE}
\SetKwInOut{Input}{Input}\SetKwInOut{Output}{Output}
\SetKwInOut{InputOutput}{Input/Output}
\SetKwProg{Fn}{Function}{:}{}
\SetKwFunction{validRegions}{validRegions}
\SetKwFunction{getMode}{getMode}
\SetKwFunction{getRegion}{getRegion}
\SetKwFunction{homeCount}{homeCount}
\SetKwFunction{ifFun}{if}

\InputOutput{Activity table $A=\langle AgentId: \mathbb{N}, Region: \mathbb{N} \cup \varnothing, \newline LocationType: \lbrace home, work, education, commercial, park \rbrace, \newline ArrivingMode: \lbrace walk, cycle, pt, car, \varnothing \rbrace \rangle$}
%\Output{Activity table $A$}
\BlankLine

\tcp{Init $anchor\_region \leftarrow \FALSE$}
\tcp{Init $primary\_mode \leftarrow \varnothing$}

%\tcp{Init $used\_pt \leftarrow \FALSE$}

% \Fn{\getMode{$region: \mathbb{N}$}}{
%   \tcp{returns the transport mode for the next region}
%   $modes \leftarrow \lbrace walk, bike , pt, car \rbrace$ \;
%   \lIf{$primary\_mode = bike$ AND $anchor\_region = \varnothing$}
%     {$modes \leftarrow \lbrace walk, bike , pt \rbrace$}
%   \lIf{$primary\_mode = car$ AND $anchor\_region = \varnothing$}
%     {$modes \leftarrow \lbrace walk, pt, car \rbrace$}
%   \lIf{$primary\_mode \neq \varnothing$ AND $anchor\_region \neq \varnothing)$}
%     {$modes \leftarrow \lbrace walk, pt \rbrace$}
%   \KwRet one of $modes$
% }
% \Fn{\validRegions{$region: \mathbb{N}$, $mode: \lbrace walk, cycle, pt, car \rbrace$, $stops: \mathbb{N}$}}{
%   \tcp{returns the set of reachable regions for a nominated region within the given number of stops using the given transport mode}
%   \KwRet $regions$
% }
% \Fn{\getRegion{$mode$, $region$, $valid\_regions$}}{
%   \tcp{returns the next region based on the current transport mode and region, restricted to a list of valid regions}
%   \KwRet $region$
% }
% \Fn{\homeCount{$row$}}{
%   \tcp{return the numbers of stops from the next home region}
%   \KwRet $n$
% }

\For{$i\leftarrow 1$ \KwTo \N}{
%  \eIf{$i=1$ OR $AgentId_{i} \neq AgentId_{i-1}$ OR $LocationType_{i} = home$}{
  % if the current location is home, but the next one isn't (avoiding home->home trips)
  \eIf{$LocationType_{i} = home$ AND $LocationType_{i+1} \neq home$}{
    $anchor\_region \leftarrow \FALSE$ \;
    $primary\_mode \leftarrow \getMode(Region_{i})$ \;
%    $used\_pt \leftarrow \ifFun(primary\_mode=pt)$ \;
    $ArrivingMode_{i+1} \leftarrow primary\_mode$ \;
    $Region_{i+1} \leftarrow getRegion(Region_{i}$, $LocationType_{i+1}$, $primary\_mode$, $homeCount(i))$ \;
    % set the arriving mode of the next home region
    $ArrivingMode_{i+\homeCount(i)} \leftarrow primary\_mode$ \;
  }{
  % \tcp{if it is possible to change modes}
  \If{ $LocationType_{i} \neq home$ AND $(homeCount(i)>2$ OR $(homeCount(i)=2$ AND $primary\_mode \in \lbrace walk,pt \rbrace))$}{
    $ArrivingMode_{i+1} \leftarrow \getMode(Region_{i}$, $primary\_mode$, $anchor\_region)$ \;
    
    % \tcp{seting anchor region and mode needed to get there}
    \If{$anchor\_region = \FALSE$ AND $primary\_mode \in \lbrace bike, car \rbrace$ AND $ArrivingMode_{i} \neq ArrivingMode_{i+1}$}{
      $Region_{i+\homeCount(i)-1} \leftarrow Region_{i}$ \;
      $ArrivingMode_{i+\homeCount(i)-1} \leftarrow ArrivingMode_{i+1}$ \;
      $anchor\_region = \TRUE$ \;
    }
    % if a walker has switched to pt, then the home region's arriving mode must be pt
    \If{$primary\_mode = walk$ AND $ArrivingMode_{i+1}=pt$}{
      $ArrivingMode_{i+\homeCount(i)} \leftarrow pt$ \;
    }
    % \tcp{selecting next region}
    \eIf{$anchor\_region = \TRUE$}{
      $Region_{i+1} \leftarrow getRegion(Region_{i}$, $LocationType_{i+1}$, $ArrivingMode_{i+1}$, $homeCount(i)-2)$ \;
    }{
      $Region_{i+1} \leftarrow getRegion(Region_{i}$, $LocationType_{i+1}$, $ArrivingMode_{i+1}$, $homeCount(i)-1)$ \;
    }
  }
 \If{$Region_{i} \neq home$ AND $homeCount(i)=2$ AND $primary\_mode \in \lbrace bike,car \rbrace$ AND $anchor\_region = \FALSE$}{
	$ArrivingMode_{i+1} \leftarrow ArrivingMode_{i}$ \;
      $Region_{i+1} \leftarrow getRegion(Region_{i}$, $LocationType_{i+1}$, $primary\_mode$, $homeCount(i)-1)$ \;
    }

}}
\caption{Algorithm for assigning transport mode and SA1 region to trips}
\label{alg:locate}
\end{algorithm}